\documentclass{article}

\usepackage[final]{corl_2019} 
\usepackage{amsmath} 
\usepackage{amssymb}  
\usepackage{bm}  
\usepackage{graphics} 
\usepackage{epsfig} 
\usepackage{wrapfig}
\usepackage{subcaption}
\usepackage[font=small,labelfont=small]{caption}
\usepackage{hyperref}
\usepackage{color}
\usepackage[table]{xcolor}
\usepackage[ruled, linesnumbered]{algorithm2e}
\usepackage{sidecap}

\usepackage[toc,page]{appendix}

\sidecaptionvpos{figure}{t}

\graphicspath{{Figures/}}

\title{Bayesian Optimization Meets Riemannian Manifolds in Robot Learning}

\author{
  No\'emie~Jaquier$^{1,2}$ \hspace{0.2cm} Leonel~Rozo$^1$ \hspace{0.2cm} Sylvain~Calinon$^2$ \hspace{0.2cm} Mathias~B\"urger$^1$ \\
  $^1$Bosch Center for Artificial Intelligence \hspace{2.1cm} $^2$Idiap Research Institute\\
  \hspace{1.4cm} Renningen, Germany \hspace{3.5cm} Martigny, Switzerland \\
  \hspace{1.cm}\texttt{name.surname@de.bosch.com} \hspace{2.3cm} \texttt{name.surname@idiap.ch} \\
  \vspace{-0.5cm}
}

\newcommand{\trsp}{\mathsf{T}}
\newcommand{\ty}[1]{{\scriptscriptstyle{\mathcal{#1}}}}

\DeclareMathOperator*{\argmax}{argmax} 
\DeclareMathOperator*{\argmin}{argmin} 

\definecolor{lightgray}{rgb}{0.9, 0.9, 0.9}

\begin{document}
\maketitle


\begin{abstract}
Bayesian optimization (BO) recently became popular in robotics to optimize control parameters and parametric policies in direct reinforcement learning due to its data efficiency and gradient-free approach. However, its performance may be seriously compromised when the parameter space is high-dimensional. A way to tackle this problem is to introduce domain knowledge into the BO framework. 
We propose to exploit the geometry of non-Euclidean parameter spaces, which often arise in robotics (e.g. orientation, stiffness matrix). Our approach, built on Riemannian manifold theory, allows BO to properly measure similarities in the parameter space through geometry-aware kernel functions and to optimize the acquisition function on the manifold as an unconstrained problem.  
We test our approach in several benchmark artificial landscapes and using a 7-DOF simulated robot to learn orientation and impedance parameters for manipulation skills.      
\end{abstract}

\keywords{Bayesian optimization, Riemannian manifolds, Robot learning} 


\section{Introduction}
When robots learn new skills or adapt their behavior to unseen conditions, their learning process needs to be safe, fast and data-efficient as the robot is a physical system interacting with the environment, making every single interaction costly. In reinforcement learning (RL) for robotics, Bayesian Optimization (BO)~\cite{Shahriari16:BayesOpt} has gained increasing interest due to its success on optimizing parametric policies in several challenging scenarios~\citep{Cully15:RobotsAsAnimal,Englert16:SuccessConstrainedBO,Marco17:LQRkernel,Rai18:BO_ATRIASbiped}. Its popularity is due to its ability to model complex noisy cost functions in a data-efficient manner, contrasting to data-hungry methods used in deep RL~\cite{Arulkumaran17_DeepRLsurvey}. However, BO performance degrades as the search space dimensionality increases, opening the door to different approaches dealing with the curse of dimensionality~\citep{Li17:HighDimBO_Dropout,Rana17:HighDimBO_elasticGP,Kirschner19:HighDimBO_1Dsubspace}. Its performance also depends on the generalization capabilities of Gaussian process (GP) models (the common surrogate model of BO), which is strongly impacted by the definition of both the mean and kernel functions.

A recent approach to improve BO performance is via domain knowledge, commonly introduced into the GP mean function~\cite{Cully15:RobotsAsAnimal} or through the design of task-specific kernels~\cite{Antonova17:DeepKernelsBO}, as detailed in~\S~\ref{sec:Relatedwork}. Nevertheless, several of these solutions are not task-agnostic, requiring new kernels as the task domain varies. A more scalable approach is to provide domain knowledge that generalizes over several tasks. In this line, we propose to provide BO with information about the geometry of the search space, a key feature often overlooked in BO applications. Geometry-awareness is particularly relevant when the parameter space is not Euclidean, which is common in robotic applications, where a variety of manifolds arise~\cite{Jaquier18:ManipTrack,Ratliff18:RiemannianMotionPolicies,Zeestraten18:PbD_Riemannian}. For example, forces and torques belong to the Euclidean manifold $\mathbb{R}^D$, stiffness, inertia and manipulability lie in the manifold of symmetric positive definite matrices $\mathcal{S}_{++}$, the special orthogonal group $SO(3)$ or the unit-sphere $\mathcal{S}^3$ are used to represent orientations, and the special Euclidean group $SE(3)$ describes robot poses. 

We hypothesize that bringing geometry-awareness into BO may improve its performance and scalability. To do so, we bring Riemannian manifold theory to Bayesian optimization (see~\S~\ref{sec:Background} for a short background). We first propose to use geometry-aware kernels allowing GP to properly measure the similarity between parameters lying on a Riemannian manifold. Second, we exploit Riemannian manifold tools to consider the geometry of the search space when optimizing the acquisition function (see~\S~\ref{sec:BO_Riemann}). These two contributions lead to a fully geometry-aware BO framework (hereinafter called GaBO) which naturally handles the constraints of parameters lying on smooth differentiable manifolds. We test GaBO with different benchmark functions and, as a proof of concept, in a set of simulated scenarios aimed at refining simple control policies for robotic manipulation (see~\S~\ref{sec:Results}). Our results show that GaBO outperforms the classical BO formulation in convergence, accuracy and scalability as the space dimensionality increases.

\vspace{-0.1cm}

\section{Related Work}
\label{sec:Relatedwork}
BO has been widely applied in diverse robotic applications, such as behavior adaptation for damaged legged robots~\citep{Cully15:RobotsAsAnimal}, controller tuning for balancing~\citep{Marco16:LQRbayesOpt}, biped locomotion~\citep{Antonova17:DeepKernelsBO,Rai18:BO_ATRIASbiped}, whole body control~\citep{Yuan19:BO_WholeBody}, physical human-robot interaction~\citep{Ghadirzadeh16:RL4pHRI,Kupcsik15:HandoverBayes}, and manipulation~\citep{Drieb17:ConstBayesOptForceTask}. A key aspect of the success of BO is the use of domain knowledge, mainly introduced into the surrogate model or the acquisition function. This prior information aims at decreasing the problem complexity and improving the convergence and accuracy of BO. This section reviews how domain knowledge has been exploited to improve BO performance.  

\citet{Cully15:RobotsAsAnimal} used simulated walking behaviors as prior knowledge that allowed a robot to quickly adapt to drastic hardware changes.~\citet{Pautrat18:BO_AutomaticPrior} extended this idea and exploited the GP mean function to introduce prior knowledge about the robot behavior, which was later used to learn new tasks using a pool of given priors. They also proposed a new acquisition function that harmonized the expected improvement and the prior model likelihood.  
~\citet{Wilson14:BO_TrajectoryKernel} used trajectory data extracted during policy execution to learn the initial state distribution, the transition and reward functions. These were used to generate Monte-Carlo estimates of policy performance that were used to define the GP mean function. The authors also introduced a Kullback-Leibler divergence kernel that used trajectory information to measure the relatedness between policies. In contrast to these works, we do not focus on imposing prior knowledge through the GP mean function but instead exploit the geometry of the parameter space to drive the BO exploration. 

\citet{Marco16:LQRbayesOpt} tuned the parameters of a pole-balancing robot controller using BO with an acquisition function maximizing the information gain. The authors emphasized that a GP with common kernels may degrade the learning outcome~\citep{Marco17:LQRkernel}, and thus kernels leveraging the controller structure may be preferred. Thus, they proposed two kernels exploiting the structure of linear quadratic regulators, which outperformed the common squared-exponential (SE) kernel.   
The shortcomings of the SE kernels were also analyzed by~\citet{Martinez17:BO_AdaptiveKernels}, who proposed a set of adaptive kernels designed to model functions from nonstationary processes. This proved useful in direct policy search where failures result in large discontinuities or flat regions while fast variations are observed around the optimal policy.
\citet{Rai18:BO_ATRIASbiped} used a gait feature transformation to generalize a range of locomotion controllers and robot morphologies. This transform reparameterized the original space of controller parameters to a 1D space, where a classical SE kernel was used. 
~\citet{Antonova17:DeepKernelsBO} used a neural network to learn a kernel using simulated bipedal locomotion patterns. This neural-network kernel could be learned from the BO cost directly or from the trajectories in a cost-agnostic manner. Unlike these works, we do not design task-specific kernels but instead exploit geometry-aware kernel functions that can be used for different tasks whose search spaces share the same geometry.
 
Another way to introduce domain knowledge into BO is via the acquisition function.~\citet{Englert16:SuccessConstrainedBO} and \citet{Drieb17:ConstBayesOptForceTask} included domain knowledge into the tuning of robot control parameters through an acquisition function using success information embedded into a GP classifier. 
\citet{Yuan19:BO_WholeBody} proposed a search space partitioning method to tackle the high dimensionality of a whole-body quadratic programming controller. The acquisition function was given only a subset of independent physically-meaningful partitions of the parameter space. In this line, our work includes domain knowledge into the acquisition function by exploiting geometry-aware optimization that handle parameters lying on Riemannian manifolds.  

The work by~\citet{Oh18:BO_Cylindrical} is one of the few in BO literature where geometry-awareness is considered. The authors applied a cylindrical transformation to the search space to overcome boundary issues when a sphere-like domain is given, and used a geometry-aware kernel. Their method was more accurate, efficient and scalable when compared to state-of-the art BO. However,~\citet{Oh18:BO_Cylindrical} did not include geometry information into the optimization of the acquisition function. As mentioned previously, our approach is fully geometry-aware as both the kernel function and the optimization of the acquisition function consider the geometry of the search space.  
  


\section{Background}
\label{sec:Background}

\subsection{Bayesian Optimization}
\label{subsec:BO}
Bayesian optimization (BO) is a sequential search algorithm aiming at finding a global maximizer (or minimizer) of an unknown objective function $f$
\begin{equation}
\bm{x}^* = \argmax_{\bm{x} \in \mathcal{X}} f(\bm{x}) ,
\end{equation}
where $\mathcal{X} \subseteq \mathbb{R}^{D_{\mathcal{X}}}$ is some design space of interest, with $D_{\mathcal{X}}$ being the dimensionality of the parameter space. The black-box function $f$ has no simple closed form, but can be observed point-wise by evaluating its value at any arbitrary query point $\bm{x}$ in the domain. This evaluation produces noise-corrupted (stochastic) outputs $y \in \mathbb{R}$ such that ${\mathbb{E}[y | f(\bm{x})] = f(\bm{x})}$, with observation noise $\sigma$.

In this setting, BO specifies a prior belief over the possible objective functions. Then, at each iteration $n$, this model is refined according to observed data $\mathcal{D}_n=\{(\bm{x}_i, y_i)\}_{i=1}^n$ via Bayesian posterior update. An acquisition function $\gamma_n : \mathcal{X} \mapsto \mathbb{R}$ is constructed to guide the search for the optimum. This function evaluates the utility of candidate points for the next evaluation of $f$; therefore, the next query point $\bm{x}_{n+1}$ is selected by maximizing $\gamma_n$, i.e. $\bm{x}_{n+1} = \argmax_{\bm{x}} \gamma_{n}(\bm{x}; \mathcal{D}_n)$.
After $N$ queries, the algorithm makes a final recommendation $\bm{x}_N$, representing its best estimate of the optimizer.

The prior and posterior of $f$ are commonly modeled using a Gaussian Process $\mathcal{GP}(\mu, k)$ with mean function $\mu :  \mathcal{X} \mapsto \mathbb{R}$ and positive-definite kernel (or covariance function) $k : \mathcal{X} \times \mathcal{X} \mapsto \mathbb{R}$.
Therefore, the function $f$ follows a Gaussian prior $f(\bm{x}) \sim \mathcal{N}(\bm{\mu}, \bm{K})$ where $\bm{\mu}_i = \mu(\bm{x}_i)$ and $\bm{K}$ is the covariance matrix with $\bm{K}_{ij} = k(\bm{x}_i, \bm{x}_j)$. With $\tilde{\bm{x}}$ representing an arbitrary test point, the random variable $f(\tilde{\bm{x}})$ conditioned on observations is also normally distributed with the following posterior mean and variance functions:
\begin{equation}
\mu_n(\tilde{\bm{x}}) = \mu(\tilde{\bm{x}}) + \bm{k}(\tilde{\bm{x}})^\trsp (\bm{K} + \sigma^2 \bm{I})^{-1}(\bm{y} - \bm{\mu})\: \text{ and }\:
\sigma_n^2(\tilde{\bm{x}}) = k(\tilde{\bm{x}},\tilde{\bm{x}}) - \bm{k}(\tilde{\bm{x}})^\trsp (\bm{K} + \sigma^2 \bm{I})^{-1} \bm{k}(\tilde{\bm{x}}), 
\end{equation}
where $\bm{k}(\tilde{\bm{x}})$ is a vector of covariance terms between $\tilde{\bm{x}}$ and the observations $\bm{x}_i$~\citep{Rasmussen06}. The posterior mean and variance evaluated at any point $\tilde{\bm{x}}$ respectively represent the model prediction and uncertainty of the objective function at $\tilde{\bm{x}}$. In BO, these functions are exploited to select the next query $\bm{x}_{n+1}$ by means of the acquisition function. The mean and kernel functions completely specify the GP and thus the model of the function $f$. The most common choice for the mean function is a constant value, while the kernel typically has the property that close points in the input space have stronger correlation than distant points. One popular kernel is the squared-exponential (SE) kernel
$k(\bm{x}_i, \bm{x}_j) = \theta \exp(-\beta d(\bm{x}_i, \bm{x}_j)^2)$,
where $d(\cdot, \cdot)$ denotes the distance between two observations and the parameters $\beta$ and $\theta$ control the horizontal and vertical scale of the function. The kernel parameters and the observation noise are usually inferred via maximum likelihood estimation (MLE).

The acquisition function balances exploitation (e.g. selecting the point with the highest posterior mean) and exploration (e.g. selecting the point with the highest posterior variance) using the information given by the posterior functions. In this paper, we use an improvement-based acquisition function, namely, expected improvement (EI)~\citep{Mockus75:EI}. 
For EI, the next query intuitively corresponds to the point where the expected improvement over the previous best observation $f^*_n$ is maximal.
\vspace{-0.2cm}

\subsection{Riemannian Manifolds}
\label{subsec:Manifolds}
In robotics, diverse type of data do not belong to a vector space and thus the use of classical Euclidean space methods for treating and analyzing these variables is inadequate. A common example is the unit quaternion, widely used to represent orientations. The quaternion has unit norm and therefore can be represented as a point on the surface of a 3-sphere. Symmetric positive definite (SPD) matrices are also widely used in robotics in the form of stiffness and inertia matrices, or manipulability ellipsoids. Both the sphere and the space of SPD matrices can be endowed with a Riemannian metric to form Riemannian manifolds. Intuitively, a Riemannian manifold $\mathcal{M}$ is a mathematical space for which each point locally resembles an Euclidean space. For each point $\bm{x}\!\in\!\mathcal{M}$, there exists a tangent space $\mathcal{T}_{\bm{x}} \mathcal{M}$ equipped with a smoothly-varying positive definite inner product called a Riemannian metric. This metric permits to define curve lengths on the manifold. These curves, called geodesics, are the generalization of straight lines on the Euclidean space to Riemannian manifolds, as they represent the minimum length curves between two points in $\mathcal{M}$ (see Fig.~\ref{Fig:Manifolds}). 

To utilize the Euclidean tangent spaces, we need mappings back and forth between $\mathcal{T}_{\bm{x}} \mathcal{M}$ and $\mathcal{M}$, which are known as exponential and logarithmic maps.
The exponential map $\text{Exp}_{\bm{x}}: \mathcal{T}_{\bm{x}} \mathcal{M}\to \mathcal{M}$ maps a point $\bm{u}$ in the tangent space of $\bm{x}$ to a point $\bm{y}$ on the manifold, so that it lies on the geodesic starting at $\bm{x}$ in the direction $\bm{u}$ and such that the geodesic distance $d_{\mathcal{M}}$ between $\bm{x}$ and $\bm{y}$ is equal to norm of the distance between $\bm{x}$ and $\bm{u}$. The inverse operation is called the logarithmic map $\text{Log}_{\bm{x}}:  \mathcal{M}\to \mathcal{T}_{\bm{x}}\mathcal{M}$.
Another useful operation over manifolds is the parallel transport $\Gamma_{\bm{x}\to\bm{y}}: \mathcal{T}_{\bm{x}}\mathcal{M}\to\mathcal{T}_{\bm{y}}\mathcal{M}$, which moves elements between tangent spaces such that the inner product between two elements in the tangent space remains constant.

In this paper, we illustrate our approach in two manifolds widely used in robotics, namely the sphere and SPD manifolds. The unit sphere $\mathcal{S}^d$ is a $d$-dimensional manifold embedded in $\mathbb{R}^{d+1}$. The tangent space $\mathcal{T}_x\mathcal{S}^d$ is the hyperplane tangent to the sphere at $\bm{x}$. The manifold of $D\!\times\!D$ SPD matrices $\mathcal{S}_{\ty{++}}^D$ can be represented as the interior of a convex cone embedded in its tangent space of symmetric matrices $\text{Sym}^{D}$. These manifolds and the foregoing operations are illustrated in Figure~\ref{Fig:Manifolds} and mathematically described in Table~\ref{Tab:Manifolds}.

\newsavebox{\spdmat}
\savebox{\spdmat}{$\left(\begin{smallmatrix}T_{11} & T_{12}\\ T_{12} & T_{22}\end{smallmatrix}\right)$}
\begin{figure}[tbp]
	\centering
	\begin{subfigure}[b]{0.49\textwidth}
		\includegraphics[width=.35\textwidth]{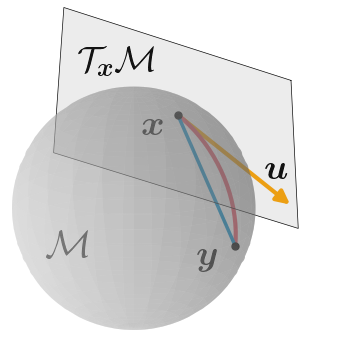}
		\includegraphics[width=.35\textwidth]{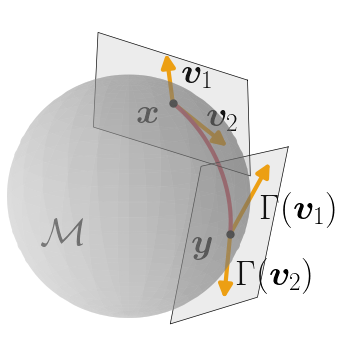}
		\includegraphics[width=.15\textwidth]{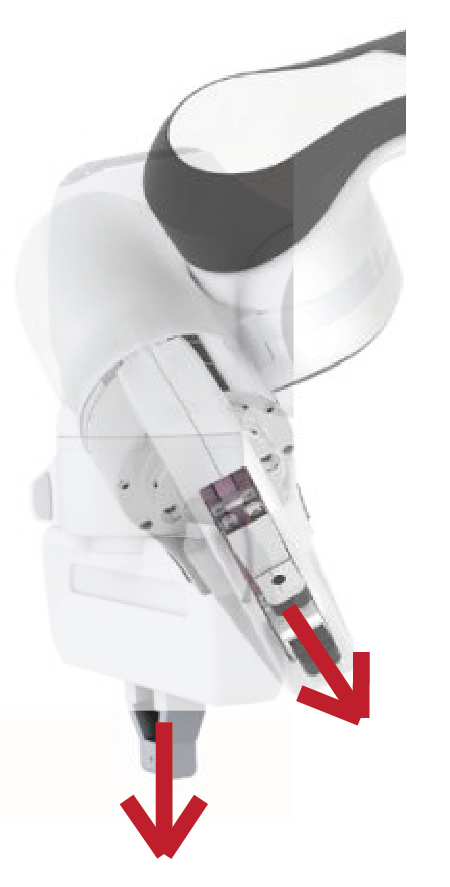}
		\caption{Sphere manifold $\mathcal{S}^2$ (incl. e.g. orientations).}
		\label{subFig:Sphere}
	\end{subfigure}
	\begin{subfigure}[b]{0.49\textwidth}
		\includegraphics[width=.35\textwidth]{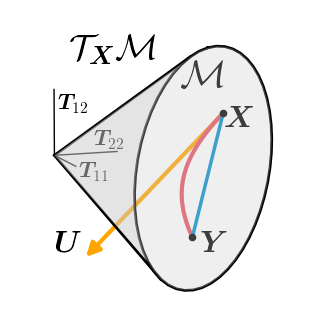}
		\includegraphics[width=.35\textwidth]{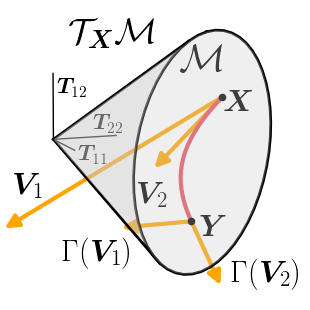}
		\includegraphics[width=.18\textwidth]{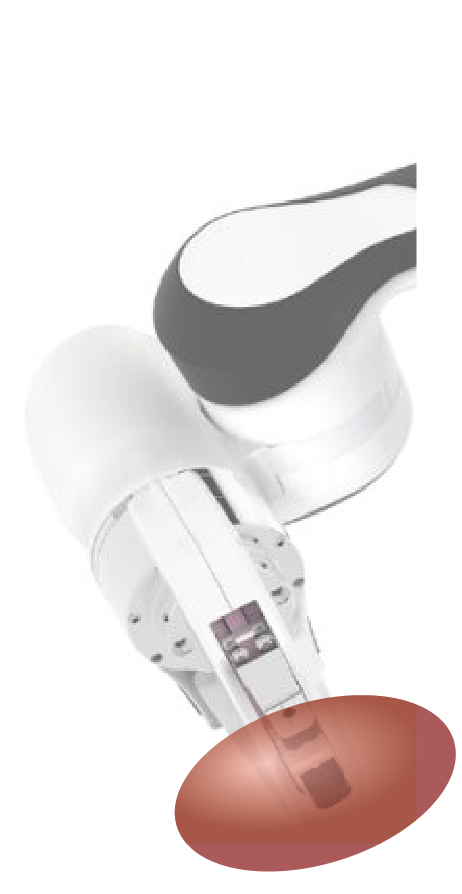}
		\caption{SPD manifold $\mathcal{S}_{\ty{++}}^2$ (incl. e.g. stiffness ellipsoids)}
		\label{subFig:SPD}
	\end{subfigure}
	\caption{(\emph{a}) Points on the surface of the sphere, such as $\bm{x}$ and $\bm{y}$ belong to the manifold. 
		(\emph{b}) One point corresponds to a matrix~\usebox{\spdmat} $\in \text{Sym}^2$ in which the manifold is embedded.
		For all the graphs, the shortest path between $\bm{x}$ and $\bm{y}$ is the geodesic represented as a red curve, which differs from the Euclidean path depicted in blue. $\bm{u}$ lies on the tangent space of $\bm{x}$ such that $\bm{u} = \text{Log}_{\bm{x}}(\bm{y})$. $\Gamma(\bm{v}_1)$, $\Gamma(\bm{v}_2)$ are the parallel transported vectors $\bm{v}_1$ and $\bm{v}_2$ from $\mathcal{T}_{\bm{x}}\mathcal{M}$ to $\mathcal{T}_{\bm{y}}\mathcal{M}$. The inner product between vectors is conserved by this operation.}
	\label{Fig:Manifolds}
	\vspace{-0.3cm}
\end{figure}

\begin{table}[t]
	\renewcommand*{\arraystretch}{1.6}
	\caption{Principal operations on the sphere $\mathcal{S}^d$ and SPD manifold $\mathcal{S}^D_{++}$ (see~\citep{Absil07,Pennec06,SraHosseini15:ConicOpt} for details).}
	\label{Tab:Manifolds}
	\scriptsize
	\begin{center}
		\begin{tabular}{c|c|c|c}
			\cellcolor{lightgray} Manifold & \cellcolor{lightgray} $d_{\mathcal{M}}(\bm{x},\bm{y}) $ & \cellcolor{lightgray} $\text{Exp}_{\bm{x}}(\bm{u})$ & \cellcolor{lightgray} $\text{Log}_{\bm{x}}(\bm{y})$ \\
			\hline
			$S^d$ & $\arccos(\bm{x}^\trsp \bm{y})$ &$\bm{x}\cos(\|\bm{u}\|) + \bm{\overline{u}}\sin(\bm{u})$ & $ d(\bm{x},\bm{y}) \, \frac{\bm{y} - \bm{x}^\trsp \bm{y} \, \bm{x}}{\|\bm{y} - \bm{x}^\trsp \bm{y} \, \bm{x}\|}$ \\
			$S^D_{++}$ & $\|\log(\bm{X}^{-\frac{1}{2}}\bm{Y}\bm{X}^{-\frac{1}{2}})\|_\text{F}$ & $\bm{X}^{\frac{1}{2}}\exp(\bm{X}^{-\frac{1}{2}}\bm{U}\bm{X}^{-\frac{1}{2}})\bm{X}^{\frac{1}{2}}$ & $\bm{X}^{\frac{1}{2}}\log(\bm{X}^{-\frac{1}{2}}\bm{Y}\bm{X}^{-\frac{1}{2}})\bm{X}^{\frac{1}{2}}$ \\
			\hline
			\cellcolor{lightgray} & \multicolumn{3}{c}{\cellcolor{lightgray}Parallel transport $\Gamma_{\bm{x}\ty{\to}\bm{y}} (\bm{v})$} \\
			\hline
			$S^d$ & \multicolumn{3}{c}{$\Big(-\bm{x}\sin(\|\bm{u}\|)\bm{\overline{u}}^{\trsp} + \bm{\overline{u}}\cos(\|\bm{u}\|)\bm{\overline{u}}^\trsp 
				+ (\bm{I}- \bm{\overline{u}}\,\bm{\overline{u}}^\trsp)\Big)\bm{v}$} with $\bm{\overline{u}}=\frac{\bm{u}}{\|\bm{u}\|}$ \\
			$S^D_{++}$ & \multicolumn{3}{c}{$\bm{A}_{\bm{X}\to\bm{Y}} \; \bm{V} \; \bm{A}_{\bm{X}\to\bm{Y}}^\trsp$ with $\bm{A}_{\bm{X}\to\bm{Y}} = \bm{Y}^{\frac{1}{2}}\bm{X}^{-\frac{1}{2}}$} \\
			\hline
		\end{tabular}
	\normalsize
	\end{center}
	\vspace{-0.4cm}
\end{table}
\normalsize 
\vspace{-0.2cm}


\section{Bayesian Optimization on Riemannian Manifolds}
\label{sec:BO_Riemann}

\vspace{-.1cm}
In this section, we present the geometry-aware BO (GaBO) framework that naturally handles the cases where the design space of parameters $\mathcal{X}$ is a Riemannian manifold or a subspace of a Riemannian manifold, i.e. $\mathcal{X} \subseteq \mathcal{M}$. To do so, we first need to model the unknown objective function $f$ with a Gaussian process adapted to manifold-valued data. This is achieved by defining geometry-aware kernels measuring the similarity of the parameters on the manifold. Moreover, the selection of the next query point $\bm{x}_{n+1}$ is achieved by optimizing the acquisition function on the manifold $\mathcal{M}$.  

\vspace{-.2cm}
\subsection{Geometry-aware Kernels}
\label{subsec:RiemannianKernels}
The choice of the kernel function is crucial for the GP as it encodes our prior about the function $f$. As our parameters $\bm{x}$ belong to a Riemannian manifold, it is relevant to include this \emph{a priori} knowledge in the choice of the kernel. A straightforward approach to adapt distance-based kernels to Riemannian manifolds is to replace the Euclidean distance $d$ by the geodesic distance $d_{\mathcal{M}}$ in the definition of the kernel. Thus, the geodesic generalization of the SE kernel is given by (see~\citep{Jayasumana15})
\begin{equation}
k(\bm{x}_i, \bm{x}_j) = \theta \exp(-\beta d_{\mathcal{M}}(\bm{x}_i, \bm{x}_j)^2).
\label{eq:ManifoldGaussianKernel}
\end{equation}
Although it has been successfully used in some applications,~\citet{Feragen15:GeodesicKernels} showed that such a kernel is valid, i.e. positive definite (PD) for all the parameters values, only if the manifold is isometric to an Euclidean space. This implies that the geodesic SE kernel is not valid for curved manifolds such as $\mathcal{S}^d$ and $\mathcal{S}_{\ty{++}}^D$. 
However, the same authors recently conjectured that there exists intervals of the lengthscale parameter $\beta > \beta_{\min}$ resulting in PD kernels~\citep{Feragen16:OpenProblem}. In this work, we follow this approach and determine experimentally the intervals of lengthscales $\beta$ for which kernel matrices are PD for the manifolds of interest. 

\begin{figure}[tbp]
	\centering
	\begin{minipage}[c]{0.6\textwidth}
		\includegraphics[width=.49\textwidth]{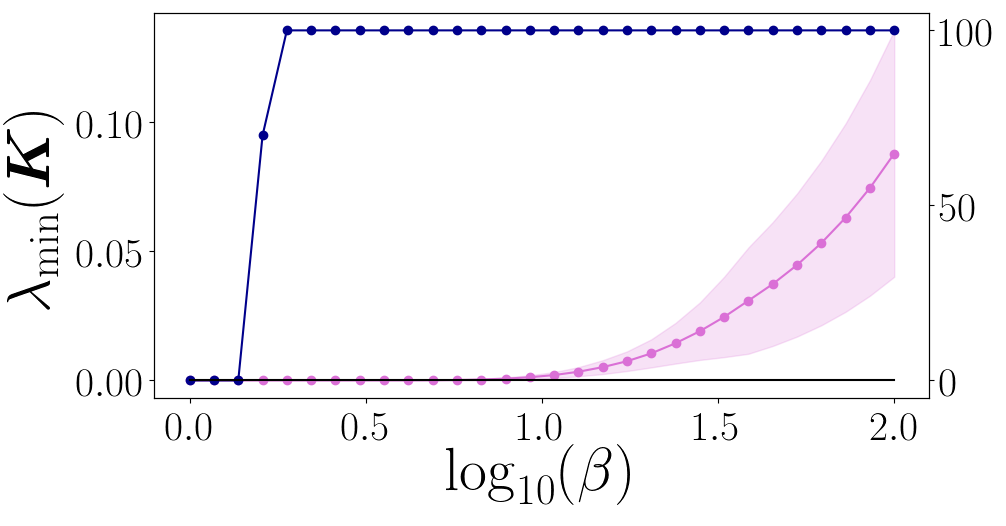}
		\includegraphics[width=.49\textwidth]{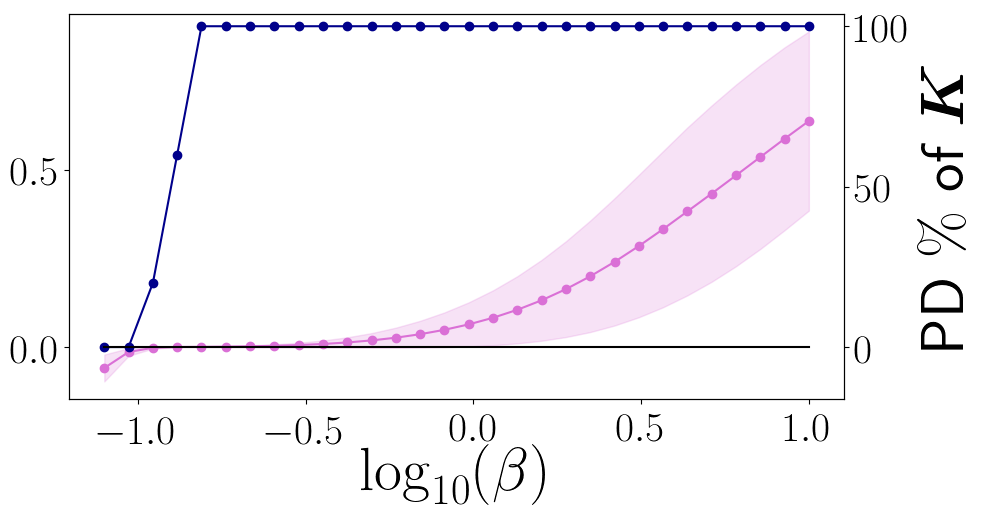}
	\end{minipage}
	\begin{minipage}[c]{0.3\textwidth}
		\centering
		\renewcommand*{\arraystretch}{1.2}
		\resizebox{\textwidth}{!}{
		\begin{tabular}{c|c|c|c|c|c}
			$\mathcal{M}$ & $\mathcal{S}^2$ & $\mathcal{S}^3$ & $\mathcal{S}^4$ & $\mathcal{S}^2_{++}$ & $\mathcal{S}^3_{++}$ \\
			\hline
			$\beta_{\min}$ & $6.5$ & $2.$ & $1.2$ & $0.6$ & $0.2$ \\
			\hline
		\end{tabular}}
	\end{minipage}
	\caption{Experimental selection of $\beta_{\min}$ in $\mathcal{S}^3$ (\emph{left} plot) and $\mathcal{S}^3_{++}$ (\emph{middle} plot). The percentage of PD geodesic SE kernel matrices $\bm{K}$ computed from $10$ different sets of $500$ samples on the manifold is depicted in blue (right axis). The corresponding distribution of minimum eigenvalue $\lambda_{\min}$ of $\bm{K}$ is depicted in purple (left axis). \emph{Right} table: selected values of $\beta_{\min}$ for different Riemannian manifolds used in this paper.}
	\label{Fig:GeodesicKernelsParams}
	\vspace{-0.3cm}
\end{figure}

In order to compute $\beta_{\min}$, we sample $500$ points from $10$ Gaussian distributions on the manifold with random mean and covariance matrix $\bm{I}$. We then compute the corresponding kernel matrix $\bm{K}_{ij}=k(\bm{x}_i, \bm{x}_j)$ for a range of $\beta$ values with $\theta=1$. We repeat this process $10$ times for each value of $\beta$ and compute the percentage of PD geodesic kernel matrices $\bm{K}$. As the minimum eigenvalue function is continuous and $\bm{K}_{\beta\to\infty} \to \bm{I}$, we fix $\beta_{\min}$ equal to the minimum value of $\beta$ for which $100\%$ of the matrices $\bm{K}$ are PD. Figure~\ref{Fig:GeodesicKernelsParams} (\emph{left}) shows the percentage of PD geodesic kernel matrices and the distribution of their minimum eigenvalue $\lambda_{\min}$ as a function of $\beta$ for $\mathcal{S}^3$ and $\mathcal{S}^3_{++}$. The values of $\beta_{\min}$ for the manifolds considered in this paper are provided in Fig.~\ref{Fig:GeodesicKernelsParams} (\emph{right}).

Other types of kernels are available for specific manifolds and may also be used in BO~\citep{Oh18:BO_Cylindrical}. For example, the geodesic Laplacian kernel is valid on spheres and hyperbolic spaces~\citep{Feragen15:GeodesicKernels}. Moreover, kernels have been specifically designed for several manifolds (see e.g.~\citep{Gong12} for the Grassmannian).

\subsection{Optimization of Acquisition Functions}
\label{subsec:Optimize_AcqFunc}
After refining the geometry-aware GP that models the unknown function $f$, the next query point $\bm{x}_{n+1}$ is selected by maximizing the acquisition function $\gamma_n$. In order to take into account the geometry of the domain $\mathcal{X} \subseteq \mathcal{M}$, we propose to optimize $\gamma_n$ using optimization techniques on manifolds. Note that the acquisition functions are not altered but their search space is modified. In this context, optimization algorithms on Riemannian manifolds constitute a powerful alternative to constrained optimization. These geometry-aware algorithms reformulate constrained problems as an unconstrained optimization on manifolds and consider the intrinsic structure of the space of interest. Also, they tend to show lower computational complexity and better numerical properties~\citep{Hu19}. 

In this paper, we use the conjugate gradient (CG) algorithm on Riemannian manifolds~\citep{Smith94:CGmanifold}, described in Algorithm~\ref{Algo:CGmanifold}, to maximize the acquisition function $\gamma_n$ (or minimize $\phi_n\!=\!-\gamma_n$), at each iteration $n$ of GaBO. The recursive process of the method involves the same steps as the Euclidean CG, namely: (\emph{i}) a line minimization along the search direction (step~\ref{AlgoStep:CGlinemin}); (\emph{ii}) the iterate update along the search direction (step~\ref{AlgoStep:CGupdate}); and (\emph{iii}) the computation of the next search direction combining the gradient of the function at the new iterate and the previous search direction (steps~\ref{AlgoStep:CGbeta}-\ref{AlgoStep:CGdir}). The differences with the Euclidean version are: (\emph{1}) as the gradient $\nabla\phi(\bm{z}_k)$, and thus the search direction $\bm{\eta}_k$, belong to the tangent space of $\bm{z}_k$, the exponential map is needed to update the iterate along the search direction; (\emph{2}) the step size is fixed by solving $\argmin_{\alpha_k} \phi_{n}\big(\text{Exp}_{\bm{z}_k}(\alpha_k\bm{\eta}_{k})\big)$ with a linesearch on the manifold (an example of linesearch algorithm on manifolds is provided in Appendix~\ref{appendix:LineSearch}); (\emph{3}) the previous search direction $\bm{\eta}_k$ has to be parallel transported to the tangent space of $\bm{z}_{k+1}$ to be combined with the gradient of the new iterate $-\nabla \phi_n(\bm{z}_{k+1})\in\mathcal{T}_{\bm{z}_{k+1}}\mathcal{M}$. Note that we presented the CG on manifold with the Hastenes-Stiefel update parameter $\beta_k^{\text{HS}}$, but other update techniques can also be extended to Riemannian manifolds~\citep{Absil07}.

Some problems may require to bound the search domain to a subspace, for example, to cope with physical limits or safety constraints in robotic systems when optimizing end-effector orientations or impedance parameters. 
In such cases, and particularly when the manifold is not a closed space, it is imperative to limit the domain of GaBO by defining boundary conditions inside the manifold. Therefore, the acquisition function is maximized over the domain $\mathcal{X} \subset \mathcal{M}$. 

While most of the literature on manifold optimization focuses on problems where the only constraint is that the solution belongs to the manifold, only few works proposed to extend constrained optimization algorithms on Riemannian manifolds. In this context, we propose to extend the bound-constrained CG method~\citep{Vollebregt14:BoundedCGmanifold} to Riemannian manifolds to cope with boundary conditions in the optimization. To do so, the steps~\ref{AlgoStep:CGupdate}-\ref{AlgoStep:CGdir} of Algorithm~\ref{Algo:CGmanifold} are updated as described in Algorithm~\ref{Algo:BoundedCGmanifold}. At each iteration, if the updated iterate $\bm{z}_{k+1}\notin\mathcal{X}$, it is projected back onto the feasible domain and the search direction is reinitialized. In $\mathcal{S}^d$, we bound the design space by setting limits on the components of $\bm{x}\in\mathcal{S}^d$. If a component is out of the limits, we fix it as equal to the closest limit and reformatted the remaining components so that $\bm{x}$ still belongs to the manifold (see Appendix~\ref{appendix:BoundsHandling} for an example). For $\mathcal{S}_{\ty{++}}^D$, we define limits on the eigenvalues $\lambda$ of the SPD matrix. If an iterate $\bm{z}_{k+1} \notin \mathcal{X}$, we project it back to the domain by reducing/increasing the maximum/minimum eigenvalue of the iterate.

Notice that some problems may require to optimize several parameters belonging to different manifolds. In this case, the domain of GaBO is a product of manifolds. 
Consequently, the kernel function corresponds to the product of the kernels on the different manifolds and the optimization of the acquisition function operates directly on the product of manifolds. 

\begin{algorithm}
	\footnotesize
	\caption{Optimization of acquisition function with CG on Riemannian manifolds}
	\label{Algo:CGmanifold}
	\KwIn{Acquisition function $\gamma_n$, initial iterate $\bm{z}_0 \in \mathcal{M}$}
	\KwOut{Next parameter point $\bm{x}_{n+1}$}
	Set $\phi_n=-\gamma_n$ as the function to minimize and $\bm{\eta}_0 = -\nabla \phi_n(\bm{z}_0)$ as search direction \;
	\For{$k=0,1\ldots,K$}{
	Compute the step size $\alpha_k$ using linesearch on Riemannian manifold \label{AlgoStep:CGlinemin}\;
	Set $\bm{z}_{k+1}=\text{Exp}_{\bm{z}_k}(\alpha_k \bm{\eta}_k)$ \label{AlgoStep:CGupdate}\;
	Compute $\beta_k^{\text{HS}} = \frac{\langle \nabla \phi_n(\bm{z}_{k+1}), \nabla \phi_n(\bm{z}_{k+1})-\nabla \phi_n(\bm{z}_{k}) \rangle_{\bm{z}_k}}{\langle \bm{\eta}_k,  \nabla \phi_n(\bm{z}_{k+1})-\nabla \phi_n(\bm{z}_{k})\rangle_{\bm{z}_k}}$ \label{AlgoStep:CGbeta}\;
	Set $\bm{\eta}_{k+1} = -\nabla \phi_n(\bm{z}_{k+1}) + \beta_k^{\text{HS}}\;\Gamma_{\bm{z}_k\to\bm{z}_{k+1}}(\bm{\eta_{k}})$ \label{AlgoStep:CGdir} \; 
	\If{the convergence criterion is reached}{break}
	}
	Set $\bm{x}_{n+1} = \bm{z}_{k+1}$       
	\normalsize
\end{algorithm}
\vspace{-0.5cm}
\LinesNumberedHidden 
\begin{algorithm}
	\footnotesize
	\caption{Update of steps~\ref{AlgoStep:CGupdate}-\ref{AlgoStep:CGdir} of Algorithm~\ref{Algo:CGmanifold} for a domain $\mathcal{X} \subset \mathcal{M}$}
	\label{Algo:BoundedCGmanifold}
	\setcounter{AlgoLine}{3}
	\ShowLn Set $\bm{z}_{k+1}=\text{Exp}_{\bm{z}_k}(\alpha_k \bm{\eta}_k)$ \;
	\eIf{$\bm{z}_{k+1} \in \mathcal{X}$}{
	Execute steps ~\ref{AlgoStep:CGbeta}-\ref{AlgoStep:CGdir} of Algorithm~\ref{Algo:CGmanifold}
	}{Project $\bm{z}_{k+1}$ to $\mathcal{X}$\;
	Set $\bm{\eta}_{k+1} = -\nabla \phi(\bm{z}_{k+1})$\;}
	\normalsize
\end{algorithm}
\vspace{-0.2cm}
\vspace{-0.2cm}
	

\section{Experiments}
\label{sec:Results}

We test GaBO using some benchmark test functions and two simulated experiments with a 7-DOF Franka Emika Panda robot. All the implementations were built on the Python libraries GPflow~\cite{GPflow17}, GPflowOpt~\cite{GPflowOpt17} and Pymanopt~\cite{PyManOpt16}. The simulated experiments were performed using Pyrobolearn~\cite{Delhaisse2019:Pyrobolearn}. All the BO implementations use EI as acquisition function and were initialized with $5$ random samples. Source codes and a video of the experiments are available at \href{https://sites.google.com/view/geometry-aware-bo}{https://sites.google.com/view/geometry-aware-bo}.

\subsection{Benchmark functions}
\label{subsec:BenchmarkFcts}
We use a couple of benchmark test functions to study the performance of GaBO in the Riemannian manifolds $\mathcal{S}^d$ and $\mathcal{S}^D_{++}$. To do so, we first project the test functions to these manifolds and then carry out the optimization by running $100$ trials with random initialization. The selection of the kernel parameters is carried out by MLE. The search domain $\mathcal{X}$ of the test functions corresponds to the complete manifold for $\mathcal{S}^d$ and to SPD matrices with eigenvalues $\lambda\in[0.001, 5]$ for $\mathcal{S}^D_{++}$.

In the case of $\mathcal{S}^d$, we compare GaBO against the classical BO which carries out all the operations in the Euclidean space (hereinafter called Euclidean BO). As the query points must belong to $\mathcal{S}^d$, i.e. $\|\bm{x}\|=1$, the maximization of the acquisition is considered a constrained problem in Euclidean BO. We minimize the Ackley function for parameter spaces of dimensionality $d = \{2, 3, 4\}$ to analyze the methods performance as the parameter dimension increases.
An example of the evolution of the surrogate model of GaBO for the Ackley function in $\mathcal{S}^2$ is shown in Appendix~\ref{appendix:SurrogateModelAckley}.
Figure~\ref{Fig:Sphere_Ackley} displays the median and the first and third quartiles of the logarithm of the simple regret along $80$ BO iterations for the three aforementioned hypersphere manifolds.
We observe that, although the performance of the two algorithms are comparable for low-dimensional hyperspheres, GaBO outperforms the Euclidean BO when the dimension increases: GaBO converges faster to a better optimizer with lower variance than Euclidean BO. This is particularly evident for the Ackley function on $\mathcal{S}^4$, where the median regret of Euclidean BO remains far from the minimum value and displays high variance, while GaBO converges to a value close to the optimum for all the trials after $70$ iterations. The fact that GaBO can be slightly slower than the Euclidean BO to converge in $\mathcal{S}^2$ may be due to the relatively high value of $\beta_{\min}$ in the kernel~\eqref{eq:ManifoldGaussianKernel} for this manifold. High $\beta_{\min}$ limits the spatial influence of each observation on the modeling of the function $f$. Thus, more observations are needed to model slowly-evolving regions of the function (see Appendix~\ref{appendix:BetaMinEffect}). A solution to this is to bound the domain of the optimization to a subspace of the manifold, which is often necessary in most real applications, and to determine the value $\beta_{\min}$ of the kernel for this subspace, resulting in a lower $\beta_{\min}$.

\begin{figure}[tbp]
	\centering
	\includegraphics[width=.224\textwidth]{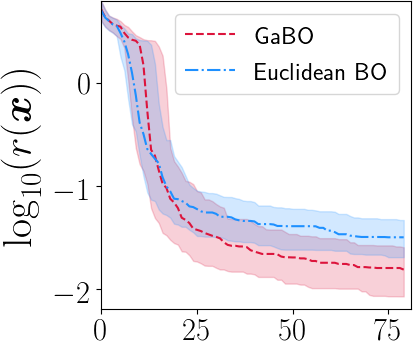}
	\includegraphics[width=0.195\textwidth]{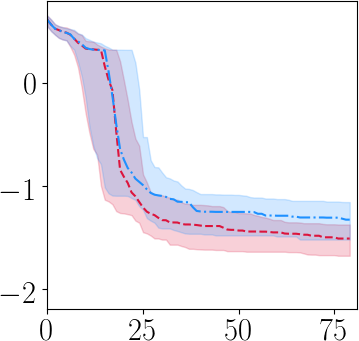}
	\includegraphics[width=0.195\textwidth]{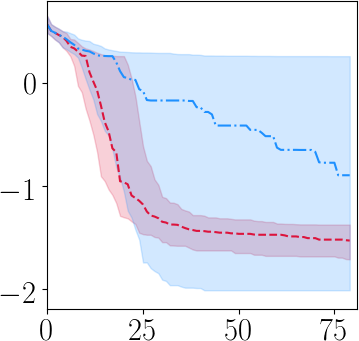}
	\includegraphics[width=.23\textwidth]{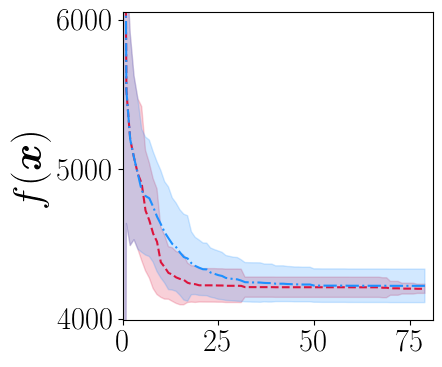}
	\caption{The three first plots show the logarithm of the simple regret in function of the BO iterations for the Ackley function in $\mathcal{S}^d$ with $d = \{2, 3, 4\}$ over 100 trials. The first and third quartiles are represented with a light tube around the median line. The last graph displays the total cost for end-effector orientation learning in $\mathcal{S}^3$ over 30 trials. The mean and one standard deviation are represented.}
	\label{Fig:Sphere_Ackley}
	\vspace{-0.3cm}
\end{figure}

\begin{figure}[tbp]
	\centering
		\includegraphics[width=.31\textwidth]{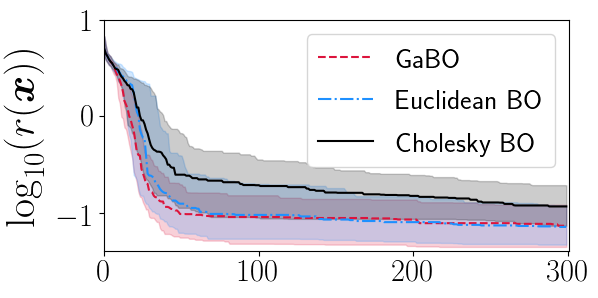}
		\includegraphics[width=.28\textwidth]{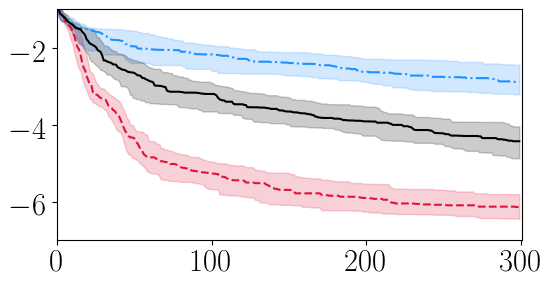}
	\caption{Logarithm of the simple regret for benchmark test functions in the Riemannian manifold of SPD matrices $\mathcal{S}^D_{++}$ over 100 trials. \emph{Left}: Ackley function in $\mathcal{S}^2_{++}$. \emph{Right}: bimodal distribution in $\mathcal{S}^2_{++}$. The first and third quartiles are represented with a light tube around the median line for the three BO methods.}
	\label{Fig:SPD_benchmark}
	\vspace{-0.4cm}
\end{figure}

Regarding the manifold $\mathcal{S}^D_{++}$, we compare our method against the Euclidean BO (augmented with the constraint $\lambda_{\min}>0$) and an alternative implementation that takes advantage of the Cholesky decomposition of an SPD matrix $\bm{A} = \bm{L}\bm{L}^{\trsp}$, so that the resulting parameter is the vectorization of the lower triangular matrix $\bm{L}$ (hereinafter called Cholesky BO). In this case, we test two functions, namely a bimodal Gaussian distribution and the Ackley function. Figure~\ref{Fig:SPD_benchmark} shows the regret for $300$ iterations of GaBO, Cholesky BO and Euclidean BO for the two test functions in $\mathcal{S}^2_{++}$. We observe that GaBO outperforms Cholesky and Euclidean BO in both cases. The computation time of GaBO is slightly higher than the Euclidean equivalent (see Appendix~\ref{appendix:ComputationTime}). However, this is not a major drawback as physical robotic experiments to collect the next value of the function $f$ will take significantly longer than the optimization process.
\vspace{-0.2cm}

\subsection{Simulated robotic experiments}
\label{subsec:SimulatedExperiments}
We here evaluate GaBO performance when looking first for the optimal orientation for a simple regulation task, and second for the optimal stiffness matrix of an impedance control policy, which is of interest for variable impedance learning approaches. For both experiments, we use a simulated 7-DOF Franka Emika Panda robot. Costs parameters values are reported in Appendix~\ref{appendix:SimulatedExpParams}.

\begin{figure}[tbp]
	\centering
	\begin{subfigure}[b]{0.23\textwidth}
		\includegraphics[width=\textwidth]{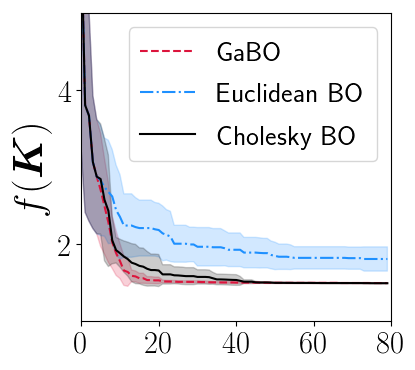}
		\caption{$f_1(\bm{K}), \bm{K}\in\mathcal{S}^2_{++}$}
		\label{subFig:SPD_stiffness2D_condNum}
	\end{subfigure}
	\begin{subfigure}[b]{0.2025\textwidth}
		\includegraphics[width=\textwidth]{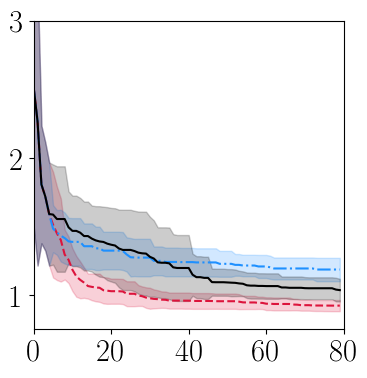}
		\caption{$f_1(\bm{K}), \bm{K}\in\mathcal{S}^3_{++}$}
		\label{subFig:SPD_stiffness3D_condNum}
	\end{subfigure}
	\begin{subfigure}[b]{0.23\textwidth}
		\includegraphics[width=\textwidth]{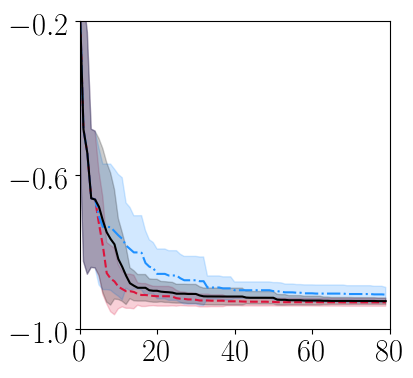}
		\caption{$f_2(\bm{K}), \bm{K}\in\mathcal{S}^2_{++}$}
		\label{subFig:SPD_stiffness2D_finalPos}
	\end{subfigure}
	\begin{subfigure}[b]{0.23\textwidth}
		\includegraphics[width=\textwidth]{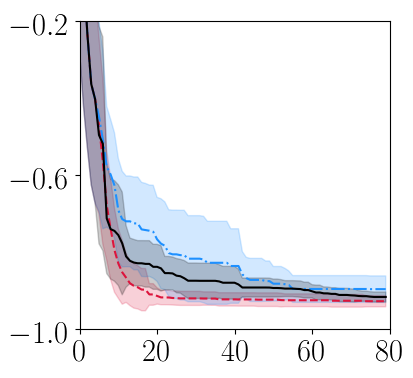}
		\caption{$f_2(\bm{K}), \bm{K}\in\mathcal{S}^3_{++}$}
		\label{subFig:SPD_stiffness3D_finalPos}
	\end{subfigure}
	\caption{Total cost for stiffness learning of a Cartesian impedance controller using direct policy search over 15 trials. The mean and one standard deviation are represented.}
	\label{Fig:SPD_stiffness}
\end{figure}
\vspace{-0.2cm}

In the first experiment, we use BO as an orientation sampler aiming at satisfying the requirements defined by a cost function. This may be useful for tasks where the orientation reference of a controller needs to be refined to improve the task execution. In this experiment, the velocity-controlled robot samples an orientation reference $\bm{x} \equiv \hat{\bm{q}}_o$ around a prior orientation $\tilde{\bm{q}}_o$, fixed by the user, with the aim of minimizing the cost function $f(\bm{q}) = w_q \| \Delta(\tilde{\bm{q}}, \bm{q}) \|^2 + w_{\bm{\tau}} \|\bm{\tau}\|^2 + w_m \text{cond}(\bm{M})$, where $\bm{q}_o$ is the current end-effector orientation, and $\text{cond}(\bm{M})$ is the condition number of the linear velocity manipulability ellipsoid. This cost function aims at minimizing the error between the prior and the current end-effector orientation with low joint torques and an isotropic manipulability ellipsoid. We run 30 trials with random initialization for each cost function. The last graph of Figure~\ref{Fig:Sphere_Ackley} shows the total cost for $80$ iterations of GaBO and Euclidean BO. We observe that GaBO converges faster to a better optimizer with a close-to-zero variance over the trials. Note that the difference between the two methods is accentuated compared to the benchmark function in $\mathcal{S}^3$.

Since direct policy search has been a successful RL approach in robotics, for our second experiment, we seek to find the optimal Cartesian stiffness of a torque-controlled 7-DOF robotic arm implementing a Cartesian control policy ${\bm{f} = \bm{K}^{\mathcal{P}}(\hat{\bm{p}}-\bm{p}) - \bm{K}^{\mathcal{D}}\dot{\bm{p}}}$, where $\bm{p}$ and $\dot{\bm{p}}$ are the linear position and velocity of the robot end-effector, $\bm{K}^{\mathcal{P}}$ and $\bm{K}^{\mathcal{D}}$ are stiffness and damping matrices, and $\bm{f}$ is the control force (transformed to desired  torques via $\bm{\tau}=\bm{J}^{\trsp}\bm{f}$). The robot task consists of tracking a desired Cartesian position $\hat{\bm{p}}$ while a constant external force $\bm{f}^e$ is applied to its end-effector. The policy parameter corresponds to the stiffness matrix, that is $\bm{x} \equiv \bm{K}^{\mathcal{P}}$. The stiffness-damping ratio is fixed as critically damped.  We tested GaBO, Euclidean BO and Cholesky BO using two different cost functions: $f_1(\bm{K}^{\mathcal{P}})= w_{\bm{p}} \|\hat{\bm{p}}-\bm{p}\|^2 + w_{d} \det(\bm{K}^{\mathcal{P}}) + w_{c}\text{cond}(\bm{K}^{\mathcal{P}})$ and $f_2(\bm{K}^{\mathcal{P}})= w_{\bm{p}} \|\hat{\bm{p}}-\bm{p}\|^2 + w_{\bm{\tau}} \|\bm{\tau}\|^2$, for $\bm{K}^{\mathcal{P}} \in \{\mathcal{S}^2_{++}, \mathcal{S}^3_{++}\}$. The cost function $f_1$ aims at accurately tracking the desired position using a low-volume isotropic stiffness matrix, while $f_2$ aims at tracking the desired position accurately with low torques. For $f_2$, a $-1$ reward was added if the desired position was reached. 
In the case of $\mathcal{S}^2_{++}$, only a $2\times 2$ submatrix of the full stiffness $\bm{K}^{\mathcal{P}}$ is optimized with BO, while the other components stay constant. Instead, all the components of the $3\times 3$ matrix $\bm{K}^{\mathcal{P}}$ are optimized in $\mathcal{S}^3_{++}$. We run 15 randomly initialized trials for each cost function. 

Figure~\ref{Fig:SPD_stiffness} shows the total cost of the stiffness learning for the two cost functions in $\mathcal{S}^2_{++}$ and $\mathcal{S}^3_{++}$. We observe that Cholesky BO tends to outperform Euclidean BO as the complexity of the cost is increased, and that GaBO outperforms the other methods for all the test cases. Moreover, while the performance of Euclidean and Cholesky BO strongly degrades as the dimensionality increases, GaBO still provides accurate and low-variance solutions.
 	


\section{Conclusion}
\label{sec:Conclusions}
\vspace{-0.1cm}
We proposed GaBO, a geometry-aware Bayesian optimization framework that exploited the geometry of the search space to properly seek optimal parameters that lie on Riemannian manifolds. To do so, we used geometry-aware kernels that allow GP to properly measure the similarity between parameters lying on a Riemannian manifold. Moreover, we exploited Riemannian manifold tools to consider the geometry of the search space when optimizing the acquisition function. GaBO provided faster convergence, better accuracy and lower solution variance when compared to geometry-unaware BO implementations. These differences were accentuated as the manifold dimensionality increased. In this line, we will evaluate GaBO in higher-dimensional manifolds. We also plan to investigate a more general projection approach to handle bound-constrained gradient methods on Riemannian manifolds. Finally, our proof-of-concept experiments open the door towards optimizing RL policies for complex robot learning scenarios where geometry-awareness may be relevant.



\clearpage
\acknowledgments{This work was carried out during a PhD Sabbatical of N. Jaquier at the Bosch Center for Artificial Intelligence (BCAI). The authors thank Andras Kupcsik and Lukas Fr\"ohlich for their useful feedback on this manuscript.}


\bibliography{References}  

\begin{appendices}
	\section{Geometry-aware kernel parameters}
\label{appendix:KernelParams}
\begin{figure}[htbp]
	\centering
	\begin{subfigure}[b]{0.31\textwidth}
		\includegraphics[width=\textwidth]{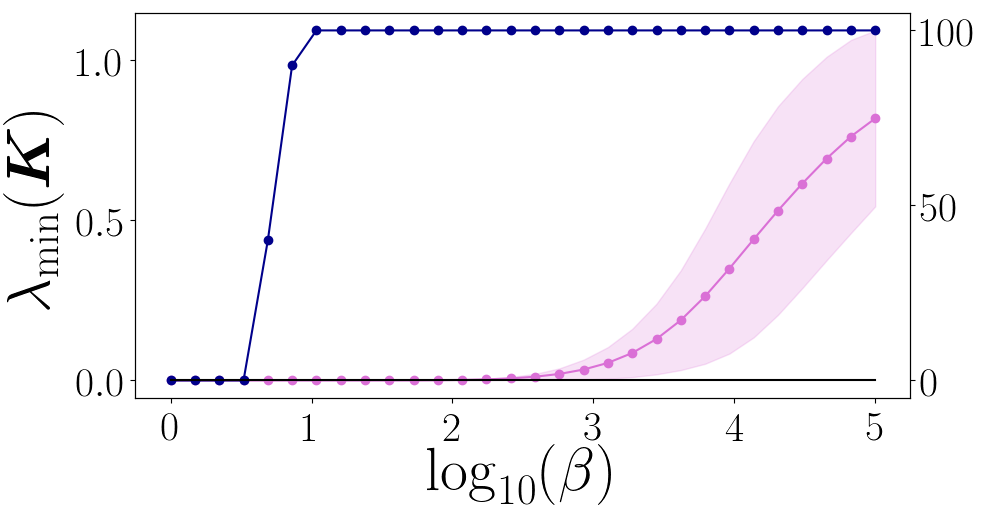}
		\caption{$\mathcal{S}^2$}
		\label{subFig:S2}
	\end{subfigure}
	\begin{subfigure}[b]{0.295\textwidth}
		\includegraphics[width=\textwidth]{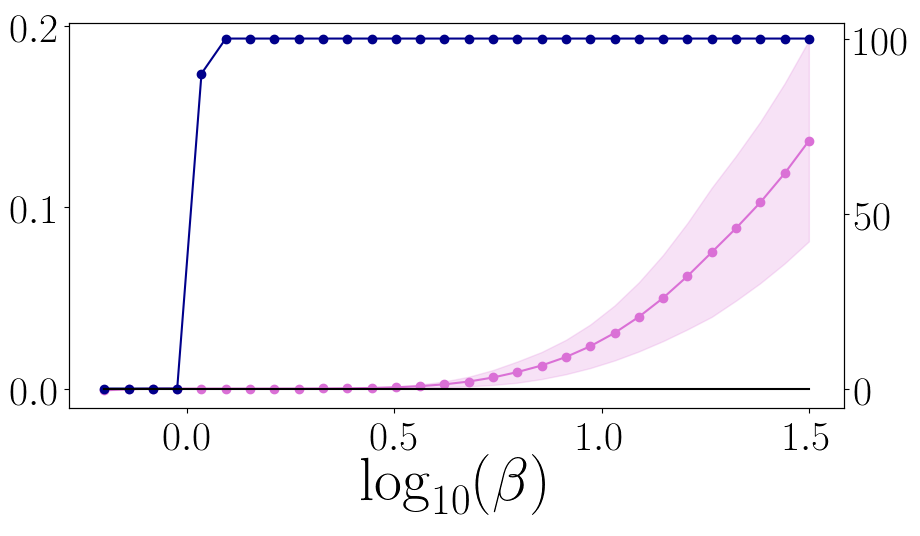}
		\caption{$\mathcal{S}^4$}
		\label{subFig:S4}
	\end{subfigure}
	\begin{subfigure}[b]{0.31\textwidth}
		\includegraphics[width=\textwidth]{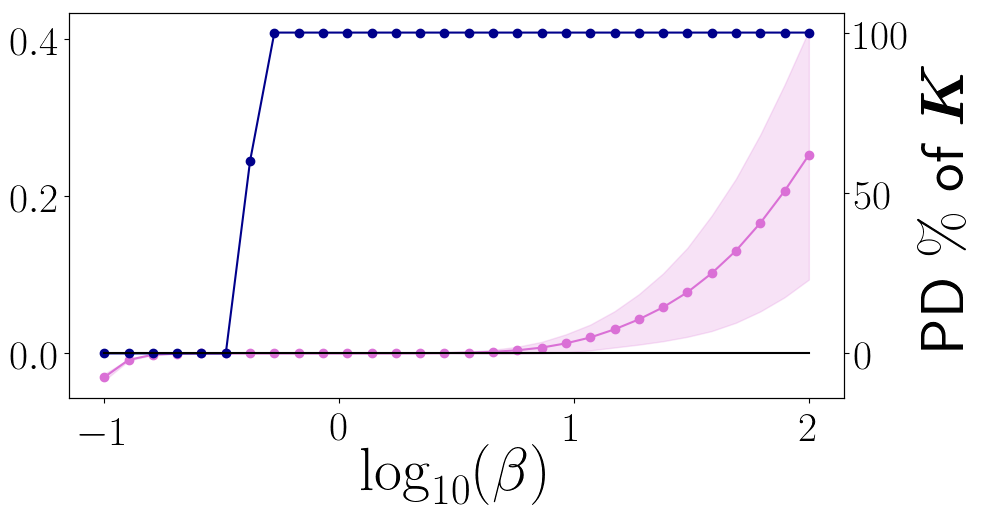}
		\caption{$\mathcal{S}^2_{++}$}
		\label{subFig:SPD3}
	\end{subfigure}
	\caption{Experimental selection of $\beta_{\min}$. The percentage of PD geodesic SE kernel matrices $\bm{K}$ computed from $10$ different sets of $500$ samples on the manifold is depicted in blue (right axis). The corresponding distribution of the minimum eigenvalue $\lambda_{\min}$ of $\bm{K}$ is depicted in purple (left axis).}
	\label{Fig:GeodesicKernelsParamsAppendix}
\end{figure}

Figure~\ref{Fig:GeodesicKernelsParamsAppendix} shows the results of the experimental selection process of $\beta_{\min}$ for the geodesic SE kernel, as introduced in Section~\ref{subsec:RiemannianKernels}, for the manifolds $\mathcal{S}^2$, $\mathcal{S}^4$ and $\mathcal{S}^2_{++}$.

\section{Influence of $\beta_{\min}$ on GaBO}
\label{appendix:BetaMinEffect}

In Section~\ref{subsec:BenchmarkFcts}, we hypothesize that the fact that GaBO can be slightly slower than the Euclidean BO to converge in $\mathcal{S}^2$ may be due to the relatively high value of $\beta_{\min}$ in the kernel~\eqref{eq:ManifoldGaussianKernel} for this manifold. In this appendix, we illustrate and analyze the effect of $\beta_{\min}$ on GaBO.

The value of $\beta$ determines the spatial influence of the observations in the Gaussian process modeling the function $f$. In other words, the value of $\beta$ controls the maximum size of the extrapolation region around one observation. 
Small $\beta$ values permit to extrapolate in a large zone around one observation and therefore are well suited to model functions that evolve slowly with only few observations. For this type of functions, where the optimum $\beta$ tends to be close to $\beta_{min}$, increasing $\beta_{min}$ may result in a slight increase of the number of observations needed to properly model the function. Therefore, GaBO may need a few more query points to converge. 

\begin{figure}[htbp]
	\centering
	\begin{subfigure}[b]{0.45\textwidth}
		\centering
		\includegraphics[width=0.42\textwidth]{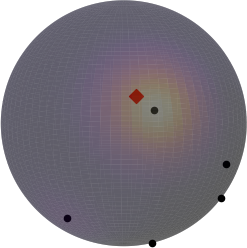}
		\includegraphics[width=0.5\textwidth]{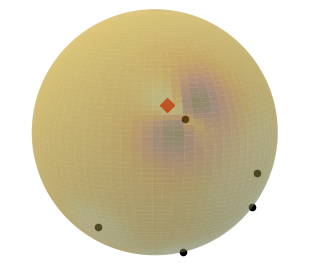}
		\caption{$\beta=6.5$}
		\label{subFig:betamin6dot5}
	\end{subfigure}
	\begin{subfigure}[b]{0.45\textwidth}
		\centering
		\includegraphics[width=0.42\textwidth]{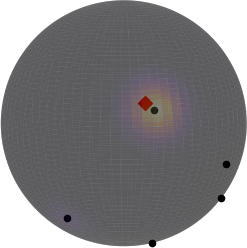}
		\includegraphics[width=0.5\textwidth]{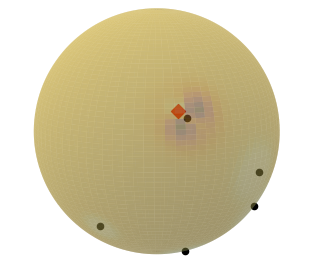}
		\caption{$\beta=25$}
		\label{subFig:betamin25}
	\end{subfigure}
	\caption{Illustration of the effect of $\beta$ on GaBO. The GP mean and corresponding acquisition function are depicted by colors ranging from yellow (low values) to dark purple (high values) on the left and right graphs, respectively. The observations and next query point are depicted by black dots and a red square.}
	\label{Fig:BetaMinEffect}
\end{figure}

Figure~\ref{Fig:BetaMinEffect} shows an example of the influence of the $\beta$ value on the choice of the next query point. Two different cases with $\beta=6.5$ and $\beta=25$ are considered. The left graphs displays the mean of a GP based on $5$ observation on the sphere with colors ranging from yellow (low values) to dark purple (high values). The corresponding acquisition function is displays on the right graphs with the same colors. The next query point, corresponding to the maximum of the acquisition function is depicted by a red square. We observe that the zone of influence of the observations is reduced for $\beta=25$ compared to $\beta=6.5$. This modifies the acquisition function, whose maximum is slightly closer to the observation for a higher value of $\beta$. Therefore, the number of query points needed to reach the optimum of the function may slightly increase if the value $\beta_{\min}$ is increased. Note that the difference between the query points obtained with $\beta=6.5$ and $\beta=25$ remains small for a consequent difference between the two tested $\beta$ values. Therefore, a slight increase of $\beta_{\min}$ will have a limited impact on the number of iterations needed by GaBO to converge.

In the case of rapidly-varying functions that are therefore better modeled by $\beta$ values significantly higher than $\beta_{min}$, a slight change on $\beta_{min}$ will not impact GaBO.

\section{Adaptive line search on Riemaniann manifolds}
\label{appendix:LineSearch}
Algorithm~\ref{Algo:LinesearchManifold} presents the adaptive line search process on Riemannian manifolds~\citep{Absil07}. The goal of this algorithm is to compute the stepsize $\alpha_k$ used in the conjugate gradient descent described in Section~\ref{subsec:Optimize_AcqFunc}. The line search aims at finding the step size that maximizes the acquisition function, i.e. it minimizes the function $\phi_n=-\gamma_n$, along the search direction $\bm{\eta}_k$. As the search direction lies on the tangent space of the iterate $\bm{z}_k$, this corresponds to solving
\begin{equation}
	\argmin_{\alpha_k} \phi_{n}\big(\text{Exp}_{\bm{z}_k}(\alpha_k\bm{\eta}_{k})\big).
\end{equation}

\begin{algorithm}
	\footnotesize
	\caption{Adaptive linesearch for CG on Riemannian manifolds}
	\label{Algo:LinesearchManifold}
	\KwIn{Function $\phi_n=-\gamma_n$, iterate $\bm{z}_k$, search direction $\bm{\eta}_k$, initial stepsize $\alpha_k$, contraction factor $c$}
	\KwOut{Final stepsize $\alpha_k$}
	\For{$j=0,1\ldots,J$}{
		Set $\bm{z} = \text{Exp}_{\bm{z}_k}(\alpha_k \eta_k)$ \; 
		\If{$\phi_n(\bm{z}) > \phi_n(\bm{z}_k) + 0.5\alpha_k \langle \nabla\phi_n(\bm{z}_k), \bm{\eta}_k \rangle_{\bm{z}_k}$}{break}
		$\alpha_k = c \alpha_k$ \;
	}
	\If{$\phi_n(\bm{z}) > \phi_n(\bm{z}_k)$}{$\alpha_k=0$}       
\end{algorithm}

\section{Example of boundary conditions handling on $\mathcal{S}^1$}
\label{appendix:BoundsHandling}
\begin{SCfigure}
	\centering
	\caption{Application of Algorithm~\ref{Algo:BoundedCGmanifold} on $\mathcal{S}^1$. The manifold $\mathcal{S}^1$ is depicted by a balck curve and the constraint $y\leq0.6$ is shown by the green dotted line. The initial point $\bm{z}$, fixed point $\bm{\tilde{z}}$ and final point $\bm{z}_{k+1}$ are depicted by red, yellow and blue dots, respectively. Arrows show the order of the projections.}
	\includegraphics[width=0.3\textwidth]{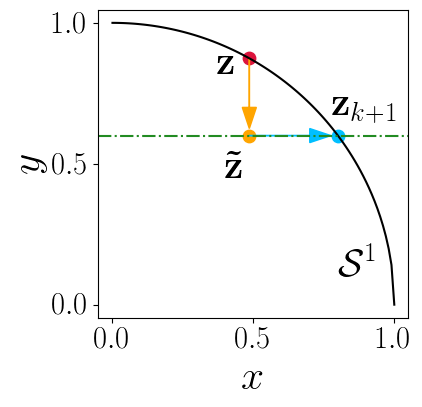}
	\label{Fig:BoundsHandlingS1}
\end{SCfigure}

Figure~\ref{Fig:BoundsHandlingS1} shows an example of the application of Algorithm~\ref{Algo:BoundedCGmanifold} on the unit circle $\mathcal{S}^1$. In this example, an acquisition function is optimized over the domain $\mathcal{S}^1$ with the constraint $y\leq0.6$. Assume that the point initially proposed by the algorithm in step~\ref{AlgoStep:CGupdate} is $\bm{z}=(x,y)^\trsp=(0.49, 0.87)^\trsp$, depicted by a red dot. This point does not satisfy the constraint $y\leq 0.6$, and needs to be projected back onto the feasible domain. As explained in Section~\ref{subsec:Optimize_AcqFunc}, we first fix the value of the $y$ component to the closest limit and obtain the point $\bm{\tilde{z}}=(0.49, 0.6)^\trsp$, depicted by a yellow dot. Note that $\bm{\tilde{z}}\notin\mathcal{S}^1$ as its norm is not equal to $1$, thus we need to project this point on the manifold. To guarantee that the bound constraints remain satisfied, we reformat only the components that were not affected by the constraints. In this example, $y$ is fixed at $0.6$ and only the component $x$ varies to obtain a point on the manifold. Therefore, we obtain the final point $\bm{z}_{k+1}=(0.8, 0.6)^\trsp\in\mathcal{X}$, depicted by a blue dot.

\section{Evolution of the surrogate model of GaBO for the Ackley function in $\mathcal{S}^2$}
\label{appendix:SurrogateModelAckley}
\begin{figure}[tbp]
	\centering
	\begin{subfigure}[b]{0.75\textwidth}
		\includegraphics[width=0.38\textwidth]{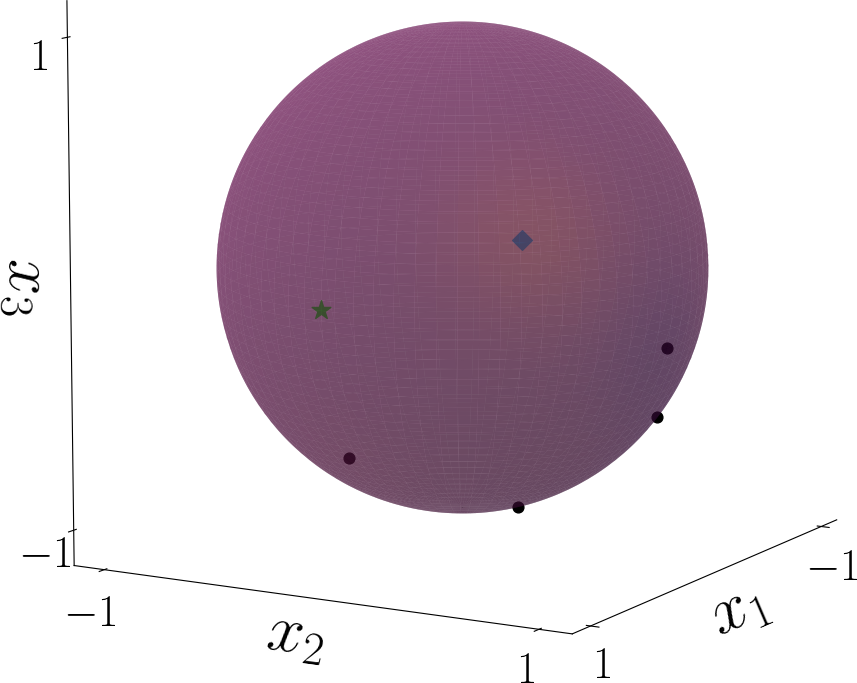}
		\includegraphics[width=.3\textwidth]{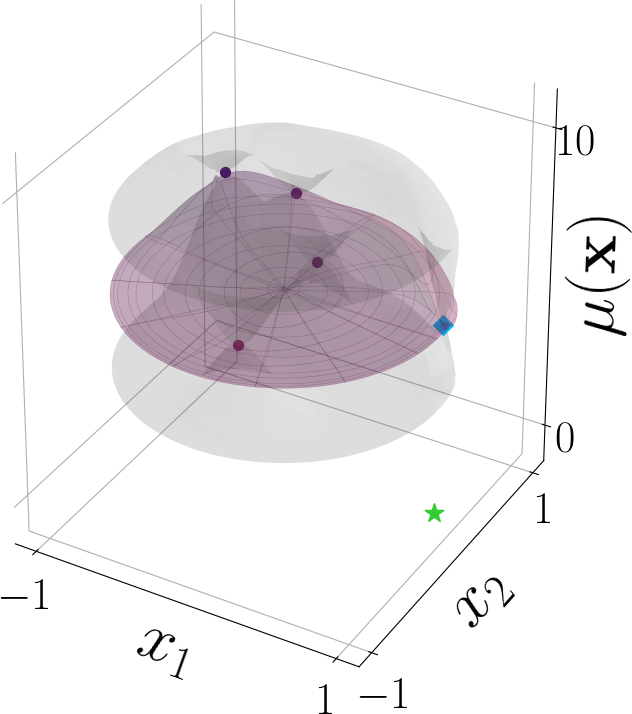}
		\includegraphics[width=.3\textwidth]{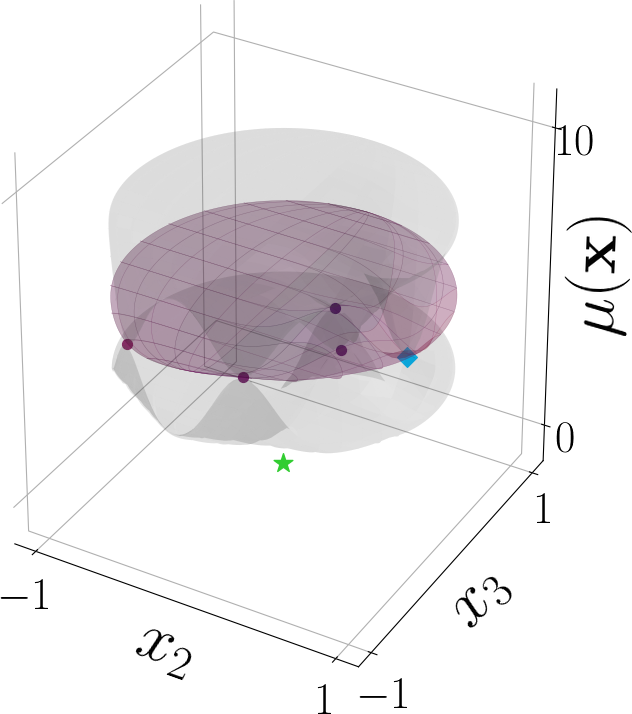}
		\caption{Initial model}
	\end{subfigure}
	\begin{subfigure}[b]{0.75\textwidth}
		\includegraphics[width=0.38\textwidth]{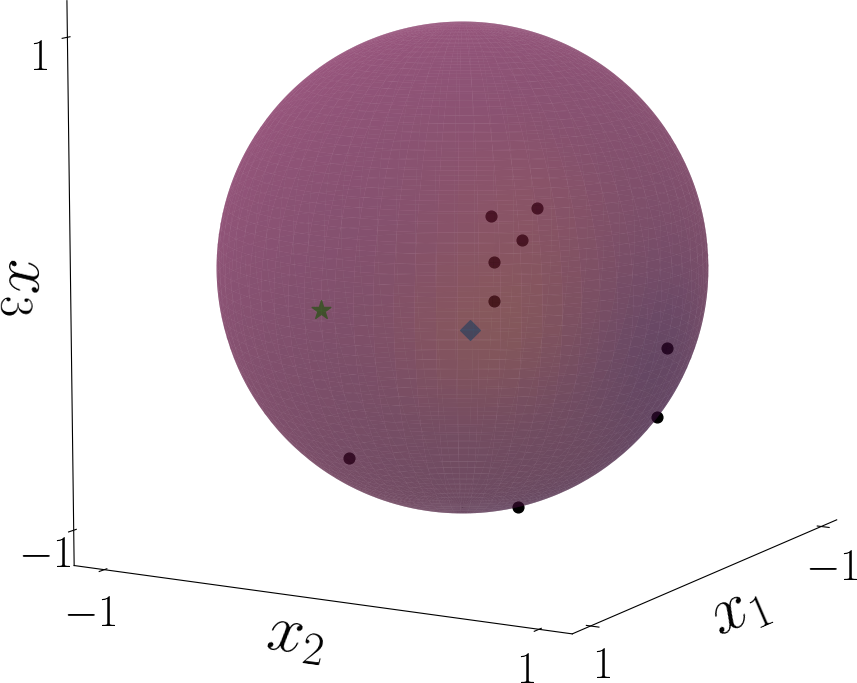}
		\includegraphics[width=.3\textwidth]{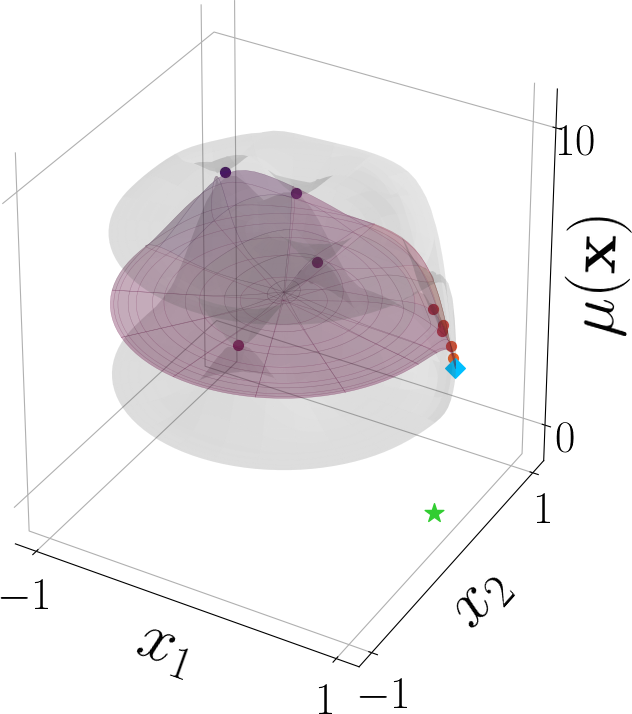}
		\includegraphics[width=0.3\textwidth]{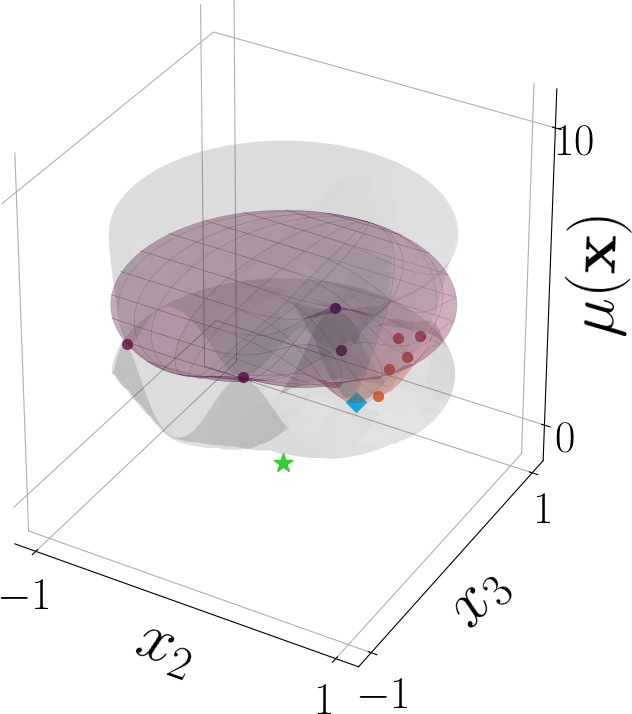}
		\caption{Iteration 5}
	\end{subfigure}
	\begin{subfigure}[b]{0.75\textwidth}
		\includegraphics[width=0.38\textwidth]{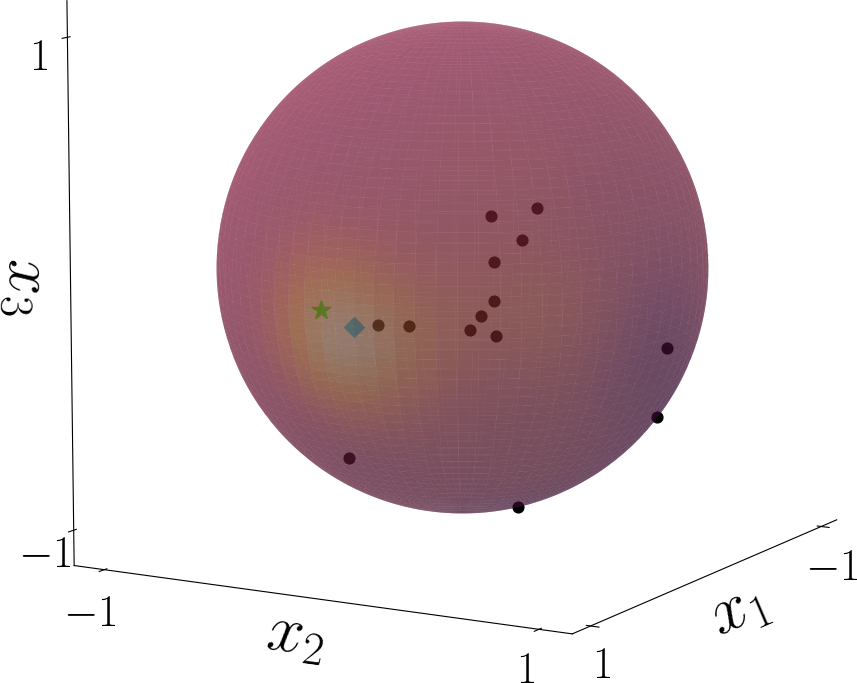}
		\includegraphics[width=.3\textwidth]{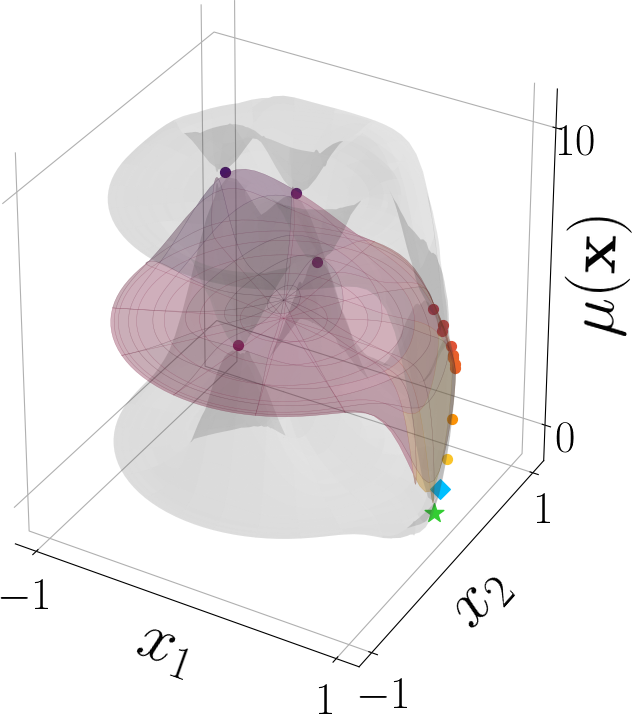}
		\includegraphics[width=0.3\textwidth]{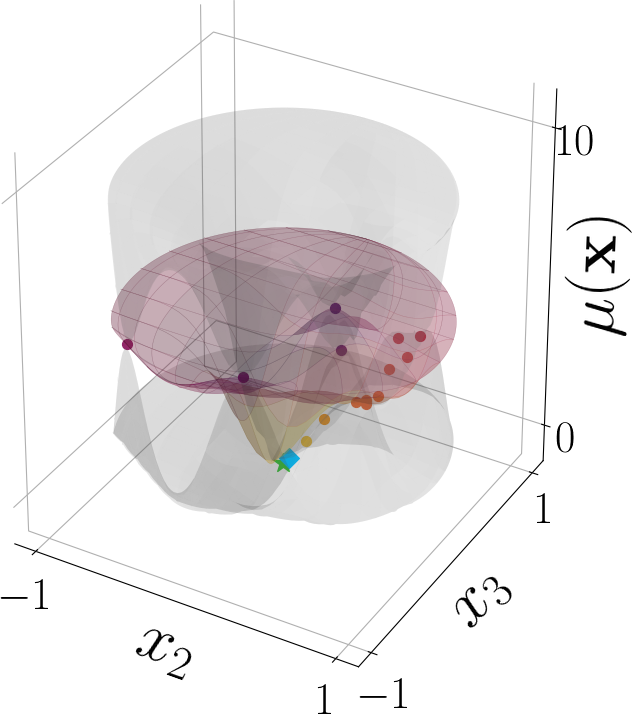}
		\caption{Iteration 10}
	\end{subfigure}
	\begin{subfigure}[b]{0.75\textwidth}
		\includegraphics[width=0.38\textwidth]{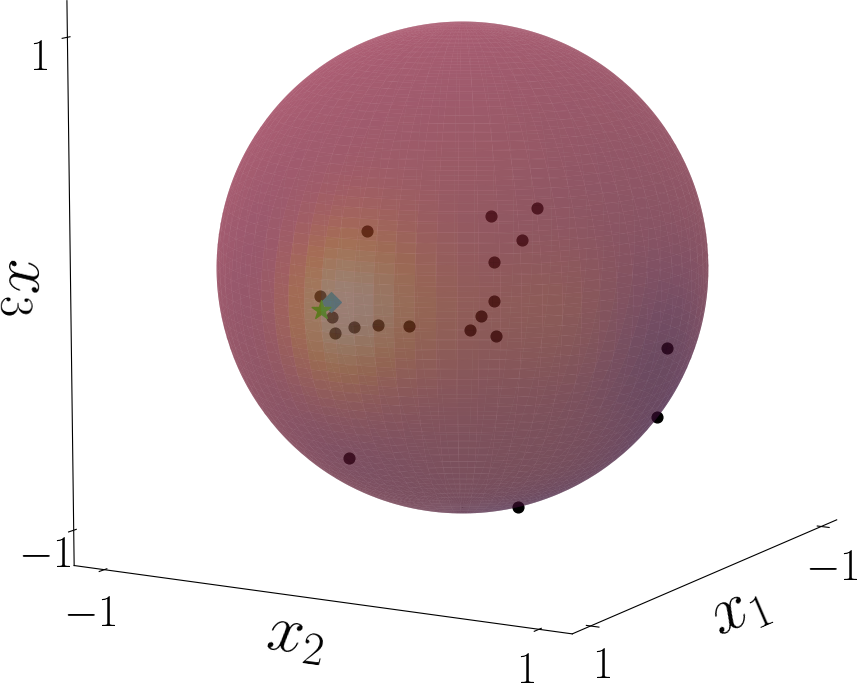}
		\includegraphics[width=.3\textwidth]{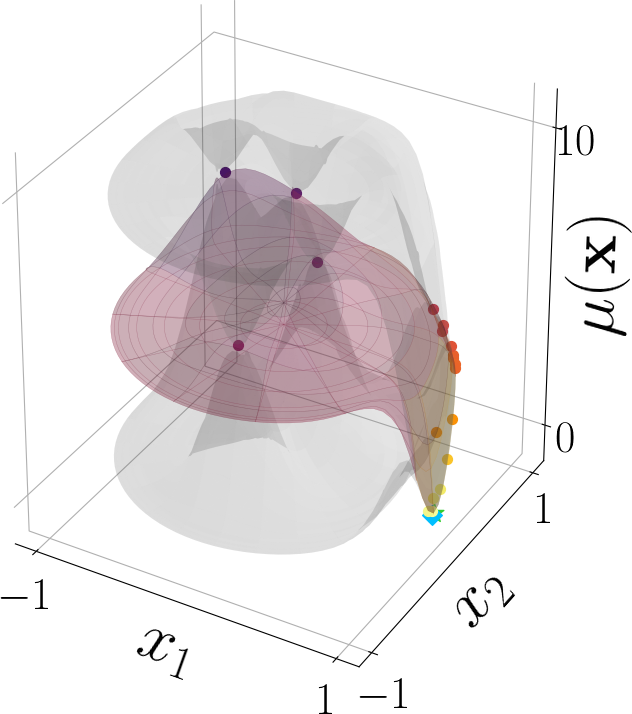}
		\includegraphics[width=0.3\textwidth]{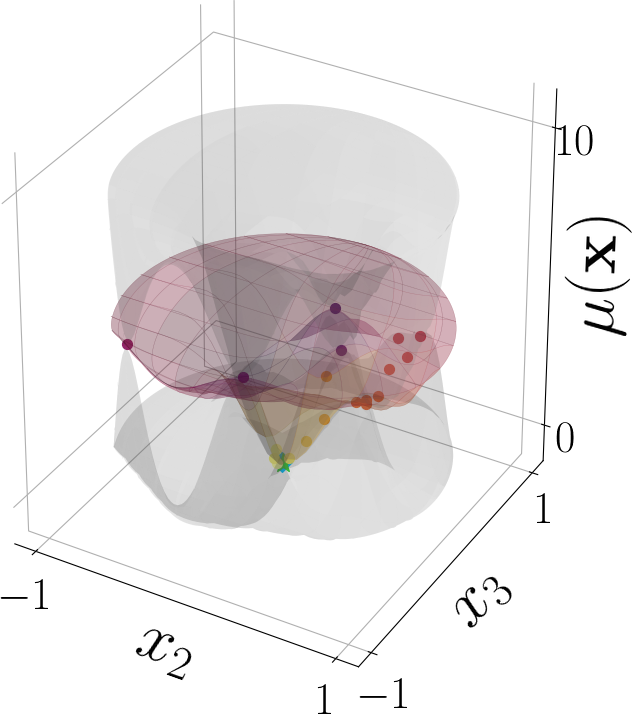}
		\caption{Iteration 15}
	\end{subfigure}
	\begin{subfigure}[b]{0.75\textwidth}
		\includegraphics[width=0.38\textwidth]{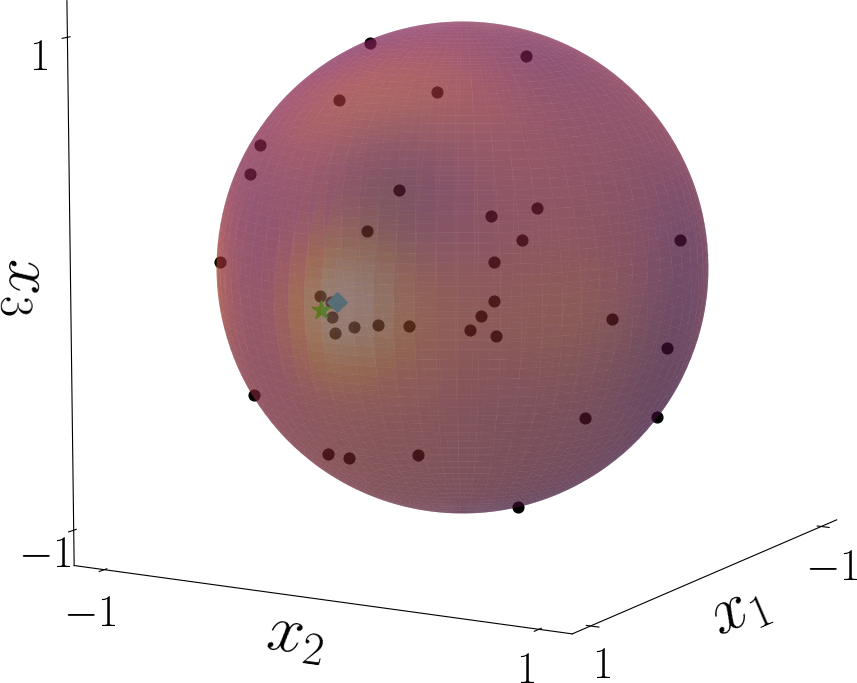}
		\includegraphics[width=.3\textwidth]{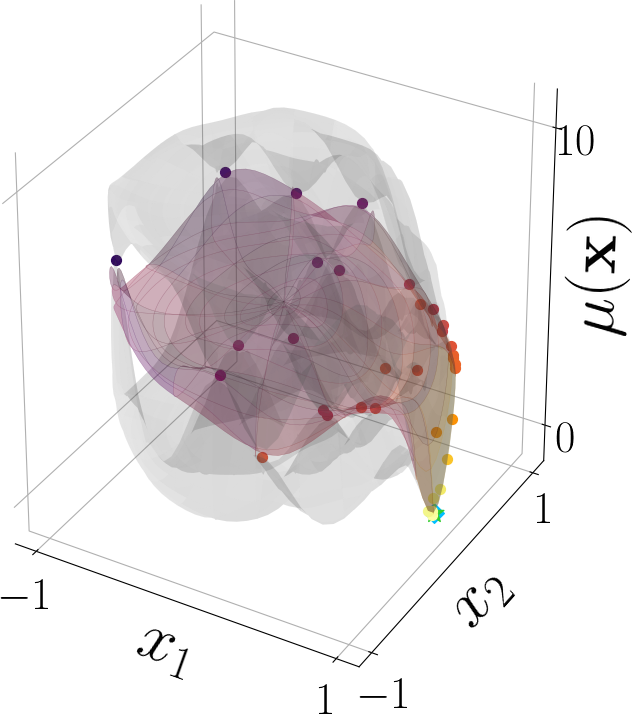}
		\includegraphics[width=0.3\textwidth]{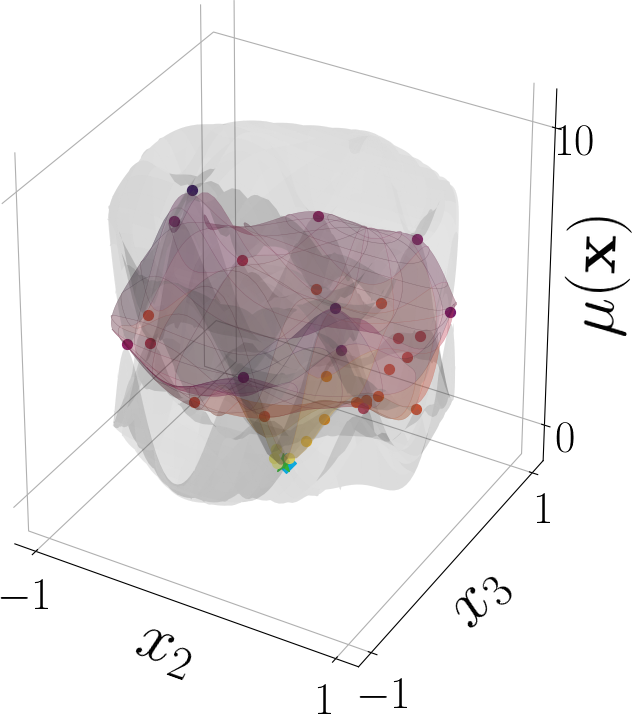}
		\caption{Iteration 30}
	\end{subfigure}
	\caption{Surrogate GP of GaBO for the minimization of the Ackley function in $\mathcal{S}^2$. The GP mean is depicted by colors ranging from yellow (low values) to dark purple (high values) and $\pm$ two standard deviations in gray (projected graphs). The global minimum and the current best guess are shown as a green star and a blue square.}
	\label{Fig:Sphere_Ackley_illustr}
	\vspace{-0.3cm}
\end{figure}

Figure~\ref{Fig:Sphere_Ackley_illustr} shows an example of the evolution of the surrogate model of GaBO for the Ackley function in $\mathcal{S}^2$. The left column displays the mean of the GP model on the sphere with colors ranging from yellow (low values) to dark purple (high values). The observed points $\bm{x}_n$ are depicted by black dots. The global minimum $\bm{x}^*=(0, 0, 0)^\trsp$ is shown as a green star and the current best guess is depicted by a blue square. The middle and right columns show the GP model on the sphere projected on the dimensions $x_1$, $x_2$ and $x_2$, $x_3$, respectively. The mean value of the GP is represented on the vertical dimension with the same colors as the left column. The variance of the model, represented with two standard deviations, is depicted by a gray envelope around the mean.

We observe that the region around the optimum of the objective function is well modeled by GP model after 10 iterations. The current best guess is also close to the optimum. After 30 iterations, the GP results in a good approximation of the objective function with a low variance in the region around the optimum value.

\section{BO computation times}
\label{appendix:ComputationTime}
Table~\ref{Tab:ComputationTime} provides the computation time per iteration of GaBO, Euclidean BO and Cholesky BO for the optimization of the Ackley function on the different manifolds used in the paper. Overall, the computation time of GaBO is slightly higher than the Euclidean equivalent. The increase observed for $\mathcal{S}^2_{++}$ when compared to $\mathcal{S}^d$ is due to the fact that the exponential and logarithmic maps are more complex in $\mathcal{S}^D_{++}$, involving matrix exponential and logarithm computations, than their equivalent on the hyperspheres. 
We do not consider the increase of the computation time as a major drawback as physical robotic experiments will take significantly longer than this time to collect the next value of the function $f$.

\begin{table}[!th]
	\renewcommand*{\arraystretch}{1.1}
	\caption{Computation time for one iteration of the different BO frameworks when optimizing the Ackley function on several manifolds. All the time values are given in seconds [s].}
	\label{Tab:ComputationTime}
	\begin{center}
		\begin{tabular}{c|c|c|c|c|}
			& $\mathcal{S}^2$ & $\mathcal{S}^3$ & $\mathcal{S}^4$ & $\mathcal{S}^2_{++}$ \\
			\hline
			GaBO & $1.55\pm 0.32$ & $1.49\pm 0.42$ & $1.94\pm0.42$& $6.08\pm5.36$ \\
			Euclidean BO & $0.48\pm0.14$& $0.49\pm0.17$ &$0.53\pm0.14$ & $0.24\pm0.17$\\
			Cholesky BO & - & - & - & $ 0.54\pm0.12 $ \\
			\hline
		\end{tabular}
	\end{center}
\end{table}
	
\section{Parameters of simulated robotic experiments}
\label{appendix:SimulatedExpParams}
Tables~\ref{Tab:OrientationCost},~\ref{Tab:StiffnessCost1} and~\ref{Tab:StiffnessCost2} provide the values of the parameters for the simulated robotic experiments presented in Section~\ref{subsec:SimulatedExperiments}. The Franka-Emika robot was initialized from a joint position $(0., 0.3, 0., -1., 0., 1.5, 0.)^\trsp$ for all the experiments.
\begin{table}[ht]
	\renewcommand*{\arraystretch}{1.3}
	\caption{Cost parameter values for the orientation sampling experiment.}
	\label{Tab:OrientationCost}
	\begin{center}
		\begin{tabular}{|c|c|c|c|}
			\hline
			$\tilde{\bm{q}}$ & $w_q$ & $w_{\bm{\tau}}$ & $ w_m$ \\
			\hline
			$(0.408, 0.408, 0, 0.816)^\trsp$ & $1$ & $10^{-4}$& $0.1$ \\
			\hline
		\end{tabular}
	\end{center}
\end{table}

\begin{table}[h!]
	\renewcommand*{\arraystretch}{1.3}
	\caption{Cost parameter values of $f_1$ in the optimal stiffness learning experiment.}
	\label{Tab:StiffnessCost1}
	\begin{center}
		\begin{tabular}{|c|c|c|c|c|c|}
			\hline
			$\hat{\bm{p}}$ [m] & $\dot{\bm{p}}$ & $w_{\bm{p}}$ & $w_{d}$ & $w_{c}$ & $\bm{f}^e$ [N]\\
			\hline
			$(0.66, -0.01, 0.69)^\trsp$ & $\bm{0}$ & $1$ & $10^{-13}$& $10^{-4}$ &  $(0, 20, -20)^\trsp$\\
			\hline
		\end{tabular}
	\end{center}
\end{table}

\begin{table}[h!]
	\renewcommand*{\arraystretch}{1.3}
	\caption{Cost parameter values of $f_2$ in the optimal stiffness learning experiment.}
	\label{Tab:StiffnessCost2}
	\begin{center}
		\begin{tabular}{|c|c|c|c|c|}
			\hline
			$\hat{\bm{p}}$ [m]& $\dot{\bm{p}}$ & $w_{\bm{p}}$  &$w_{\bm{\tau}}$ & $\bm{f}^e$ [N] \\
			\hline
			$(0.5, -0.4, 0.75)^\trsp$ & $\bm{0}$ & $1$ & $10^{-5}$ &  $(0, 20, -20)^\trsp$\\
			\hline
		\end{tabular}
	\end{center}
\end{table} 
\end{appendices}
\clearpage

\end{document}